\documentclass[10pt,twocolumn,letterpaper]{article}

\usepackage[algorithms]{wacv}      

\definecolor{wacvblue}{rgb}{0.21,0.49,0.74}
\usepackage[pagebackref,breaklinks,colorlinks,allcolors=wacvblue]{hyperref}

\usepackage{array,multirow,graphicx}
\usepackage{tikz}
\usetikzlibrary{positioning,arrows.meta,calc}
\usepackage{amssymb}
\usepackage{graphicx}
\usepackage{booktabs}
\usepackage{wrapfig}

\usepackage{comment}
\usepackage{booktabs}
\usepackage{multirow}
\usepackage{colortbl}
\usepackage{bbm}
\usepackage{arydshln}
\usepackage{caption}
\usepackage{pifont}
\usepackage[accsupp]{axessibility}  
\usepackage{hyperref}
\usepackage{orcidlink}

\title{Depth-guided Geometric Adaptation for Few-shot Image Classification}

\author{
Zhiqi Miao$^{1}$\quad
Estefanía Talavera$^{2}$\quad
Weichuan Zhang$^{3}$\quad
Jiapan Guo$^{1}$
\\[4pt]
$^{1}$University of Groningen\quad
$^{2}$University of Twente\quad
$^{3}$Shanxi University of Science and Technology
\\[4pt]
{\ttfamily\small
z.miao@rug.nl,\ e.talaveramartinez@utwente.nl,\ zwc2003@163.com,\ j.guo@rug.nl}
}

\begin{document}
\maketitle
\begin{abstract}
Few-shot image classification (FSIC) frequently suffers from visual ambiguity, where background clutter and RGB appearance similarities hinder discriminative feature matching in data-scarce settings. To mitigate this, we propose DAFIC, a geometry-guided framework that incorporates monocular depth into representation learning, episodic adaptation, and dense support-query matching. DAFIC first learns geometry-aware token representations through RGB-D self-supervised pre-training. Subsequently, the Depth-guided Geometric Adaptation (DGA) module uses depth-derived geometry to refine patch representations with geometry-enriched class context while suppressing background interference. Spatial Bidirectional Patch Matching (SBiM) then performs bidirectional dense matching on the refined features, reducing sensitivity to spurious correspondences and background clutter. Extensive experiments on four benchmarks show that DAFIC achieves strong performance against existing methods, with ablation and robustness analyses supporting the effectiveness of coupling geometry-conditioned episodic adaptation with bidirectional matching for few-shot visual recognition. Code is available at \href{https://github.com/MiaoMeow98/DAFIC}{DAFIC}.
\end{abstract}
\vspace{-3mm}
    
\section{Introduction}

Few-Shot Learning (FSL) aims to achieve human-like generalization from limited examples. Within this paradigm, Few-Shot Image Classification (FSIC) needs to recognize novel classes during inference that are disjoint from those seen during training, given only one or a few labeled samples per class \cite{wang2020generalizing,li2023libfewshot}. A critical challenge in this setting is that images in real-world scenarios and common benchmarks (e.g., ImageNet \cite{russakovsky2015imagenet}) often contain multiple entities but are assigned a single image-level label. Conventional large-scale models mitigate this by relying on abundant training data to implicitly learn to isolate class-relevant entities \cite{hiller2022Fewtrue, hao2023CPEA}. However, this strategy often fails in FSIC, where extreme data scarcity prevents the model from inherently suppressing background clutter and ambiguity.

\begin{figure}[t]
    \centering
    \includegraphics[width=0.99\columnwidth, trim=30 25mm 0mm 20mm, clip]{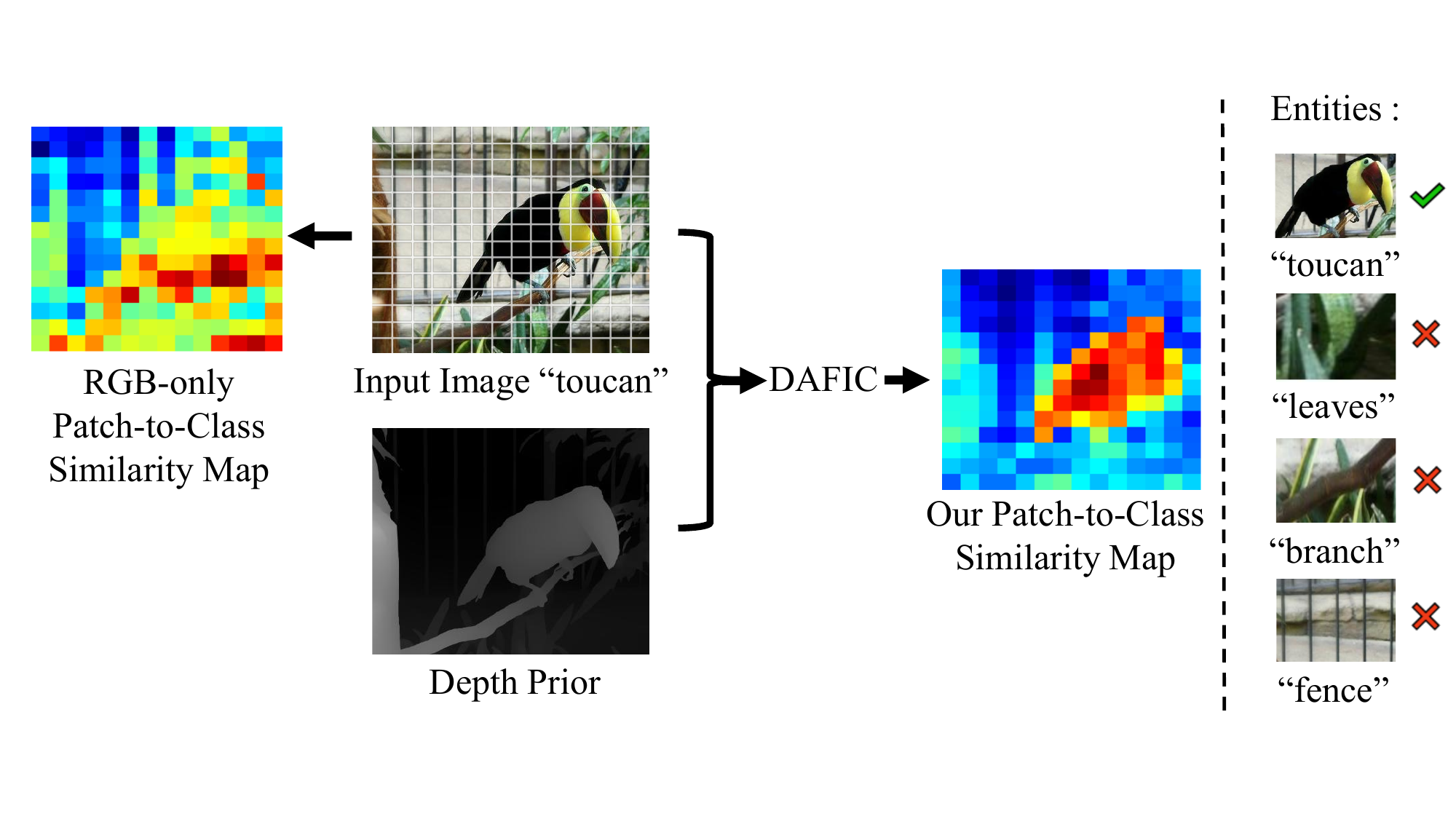}
    \caption{Patch-to-class similarity maps on a MiniImageNet~\cite{vinyals2016MatchingNet} example. In comparison, DAFIC leverages estimated depth priors to produce more target-focused similarities and suppress responses to background distractors (e.g., leaves/branch/fence).}
    \label{motivation}
    \vspace{-5mm}
\end{figure}

Motivated by this challenge, many recent metric-based FSIC approaches \cite{doersch2020CTX,hiller2022Fewtrue,hao2023CPEA,jiang2024CPES,cumplido2025slot,cumplido2025stochastic} have shifted from image-level to patch-level representations. This transition facilitates fine-grained local comparisons across images. While pioneering works refine patch embeddings to highlight class-discriminative regions via online optimization \cite{hiller2022Fewtrue}, global context adaptation \cite{hao2023CPEA,jiang2024CPES,cumplido2025slot}, or masked knowledge distillation \cite{lin2023SMKD}, they share a common limitation: they primarily rely on RGB appearances (e.g., texture, color, and illumination), without explicit geometric cues (see Fig. \ref{motivation}). This reliance leads to three major issues in cluttered scenes: (1) appearance shifts create false similarities between target entities and background distractors; (2) lacking geometric constraints, feature aggregation becomes noisy as it mixes salient object regions with surrounding clutter; and (3) the model struggles to align local patch features with global semantic context. Crucially, these effects are further amplified in episodic FSIC, resulting in class-inconsistent matching.

Furthermore, even with well-refined patch embeddings, establishing reliable image-to-class matching remains challenging. Existing methods typically employ either global spatial aggregation \cite{hiller2022Fewtrue, hao2023CPEA, lin2023SMKD}, which discards critical patch-to-patch local correspondence, or flattening dense similarity matrices into fixed parametric heads (e.g., linear layers or MLPs) \cite{hao2023CPEA, jiang2024CPES, cumplido2025slot}, which destroys the 2D layout. While non-parametric dense matching offers an intuitive alternative, it faces practical challenges in cluttered scenes. Relying solely on unidirectional matching (e.g., query-to-support) \cite{Li2019DN4} can lead to redundant assignments, where multiple query patches repeatedly match to the most salient support patch, ignoring the overall object structure. In contrast, symmetric bidirectional matching encourages better coverage and structural consistency, but it can be sensitive to unaligned background clutter; when background patches fail to find meaningful matches, they often suppress the overall image similarity score.

In this paper, we present \textbf{DAFIC}, a framework that addresses these challenges by systematically incorporating monocular geometry into few-shot image classification. Our core insight is that monocular depth can be exploited as a geometric prior through learnable adaptation, reducing patch-level semantic ambiguity and improving support-query correspondence under background clutter. Specifically, DAFIC first employs an RGB-D self-supervised pre-training scheme to learn geometry-aware token embeddings. We then introduce the \textbf{Depth-guided Geometric Adaptation (DGA)} module, which transforms depth maps into geometry-salient patch attention and further aligns patch representations with a geometry-enriched class embedding. Importantly, DGA encourages geometry-salient patches to preserve fine-grained local cues while injecting geometry-enriched class-level context, and steers low-attention (often background) regions toward more consistent global descriptors. Building on these refined embeddings, we adopt a non-parametric Spatial Bidirectional Patch Matching (SBiM) classifier to perform bidirectional dense patch matching, improving correspondence coverage and reducing sensitivity to spurious matches. Extensive experiments on four benchmarks demonstrate that  DAFIC achieves strong performance against existing methods, with ablations and robustness analyses validating the designs.

\section{Related work}

\subsection{Few-shot Image Classification}
Few-shot Image Classification (FSIC) aims to recognize novel query samples given only a few support samples. Existing FSIC approaches broadly fall into three paradigms: optimization-based methods \cite{finn2017model,nichol2018first,zintgraf2019fast,rusu2018meta,bertinetto2018CifarFS,Oh2021BOILTR} that meta-learn an initialization and an update rule for rapid adaptation, transfer learning-based methods \cite{chen2019closer,liu2020negative,mangla2020charting,rajasegaran2020self,zheng2023diffkendall} that rely on powerful base-class pre-training followed by fine-tuning (or adaptation) on novel classes, and metric-based methods \cite{snell2017ProtoNet,vinyals2016MatchingNet,Li2019DN4,li2019CTM,zhang2022DeepEMD}. Among these, metric-based learning has emerged as an intuitive and effective solution by comparing query and support samples in a high-dimensional representation space. Recent metric-based frameworks have achieved noticeable success by integrating the powerful pre-training and episodic meta-learning strategies from the other two paradigms \cite{ye2020FEAT,hou2019cross,doersch2020CTX,wertheimer2021FRN,zhang2022DeepEMD,cheng2023FGFL, xie2022DeepBDC}.

Building upon these, the focus of metric-based FSIC has increasingly shifted from image-level representations towards leveraging Vision Transformers (ViTs) \cite{dosovitskiy2020ViT} to establish fine-grained, patch-level correspondences \cite{doersch2020CTX, hiller2022Fewtrue, hao2023CPEA, lin2023SMKD, cumplido2025slot, jiang2024CPES, cumplido2025stochastic}. These approaches typically employ advanced pre-training paradigms followed by various feature adaptation strategies to emphasize semantically relevant regions. For instance, FewTURE \cite{hiller2022Fewtrue} determines patch importance via online optimization to re-weight correspondences. SMKD \cite{lin2023SMKD} employs supervised masked knowledge distillation to force patch embeddings to capture discriminative local semantics. Another line of work explicitly exploits class-aware embeddings: CPEA \cite{hao2023CPEA} uniformly broadcasts a global class embedding to all patches, while CPES \cite{jiang2024CPES} specifically selects a subset (e.g., top-$k$) of class-aware patch features. More recently, attention-driven filtering mechanisms have been explored, such as SAFF \cite{cumplido2025slot}, which iteratively updates attention via slot-attention, and SPFF \cite{cumplido2025stochastic}, which employs multinomial sampling to perform stochastic patch filtering.

Despite these advances in refining patch representations, a common limitation is that they primarily rely on 2D RGB appearance and a lack of explicit depth-derived geometric guidance. This makes them vulnerable to appearance shifts and background clutter. In comparison, DAFIC introduces monocular depth priors to learn geometry-salient patch embeddings and suppress distractors. Beyond feature refinement, prior FSIC methods explore different dense patch-matching methods, such as nearest-neighbor aggregation~\cite{Li2019DN4}, optimal transport distance~\cite{zhang2022DeepEMD}, and cross-attention mechanisms~\cite{doersch2020CTX}. We adopt a lightweight symmetric patch-matching strategy, inspired by set matching \cite{Dekel_2015_CVPR,Hausler_2021_CVPR} and few-shot action recognition \cite{HyRSM_cvpr2022,Molo_cvpr2023}, where geometry-aware representation and adaptation complement dense matching to improve robustness in cluttered scenes.

\subsection{Geometric Priors and Depth Estimation} 

\textbf{Depth for Scene Understanding.} Depth serves as a critical bridge between 2D RGB appearances and the 3D scene geometry. This has motivated the rapid evolution of RGB-D architectures in general vision tasks, ranging from dual-branch networks to unified frameworks~\cite{wang2022multimodal,zhang2023cmx,yin2024dformer}. These models increasingly integrate depth priors into core components (e.g., self-attention mechanisms) to enhance geometry-enriched representation learning. Furthermore, recent breakthroughs in zero-shot monocular depth estimation (MDE), such as generative-based Marigold \cite{Marigold2024} and discriminative-based Depth Anything \cite{yang2024depth1, yang2024depth2}, have made monocular depth a practical source of geometric cues. In particular, Depth Anything V2 leverages large-scale data (real and synthetic) to enable direct depth regression with strong open-world generalization, providing a practical source of geometric cues for downstream tasks.

\textbf{Geometric Cues in FSIC.} Despite its success in general vision, the potential of depth in FSIC remains relatively underexplored. Currently, a line of approaches rely on prompt-engineered language priors with Vision-Language Models (VLMs) \cite{zhang2024simple,li2024fewvs,liu2025envisioning} or frequency-domain cues \cite{chen2021sPlusf,cheng2023FGFL,zhou2024metaFreq}. However, these methods present their limitations: language priors often depend heavily on massive pre-trained backbones and intricate prompt engineering, while frequency cues discard spatial structure due to frequency-domain processing and can be sensitive to texture shifts. In a related setting, while prior work \cite{fu2020depthfsl} explored depth for few-shot video recognition, it utilized depth as a parallel encoding stream with extra computational overhead.

In contrast, we propose a vision-based and structure-aware approach that prioritizes spatial geometric understanding over high-level semantic or frequency-based cues. This shift to depth priors offers two pivotal advantages for FSIC: (1) depth provides an annotation-free, relatively stable signal that captures explicit structural layouts (e.g., boundaries, layering, and occlusion), which can help alleviate RGB appearance ambiguity; and (2) these geometric cues can serve as an initial spatial prior, which can be optimized via episodic training to generate geometry-salient attention for isolating target entities. Therefore, our proposed DAFIC explicitly augments episodic learning with these geometric structures to obtain target-entity salient representations for few-shot similarity measurement, providing an effective alternative to existing paradigms.

\begin{figure*}[t]
    \centering
    \includegraphics[width=0.92\textwidth]{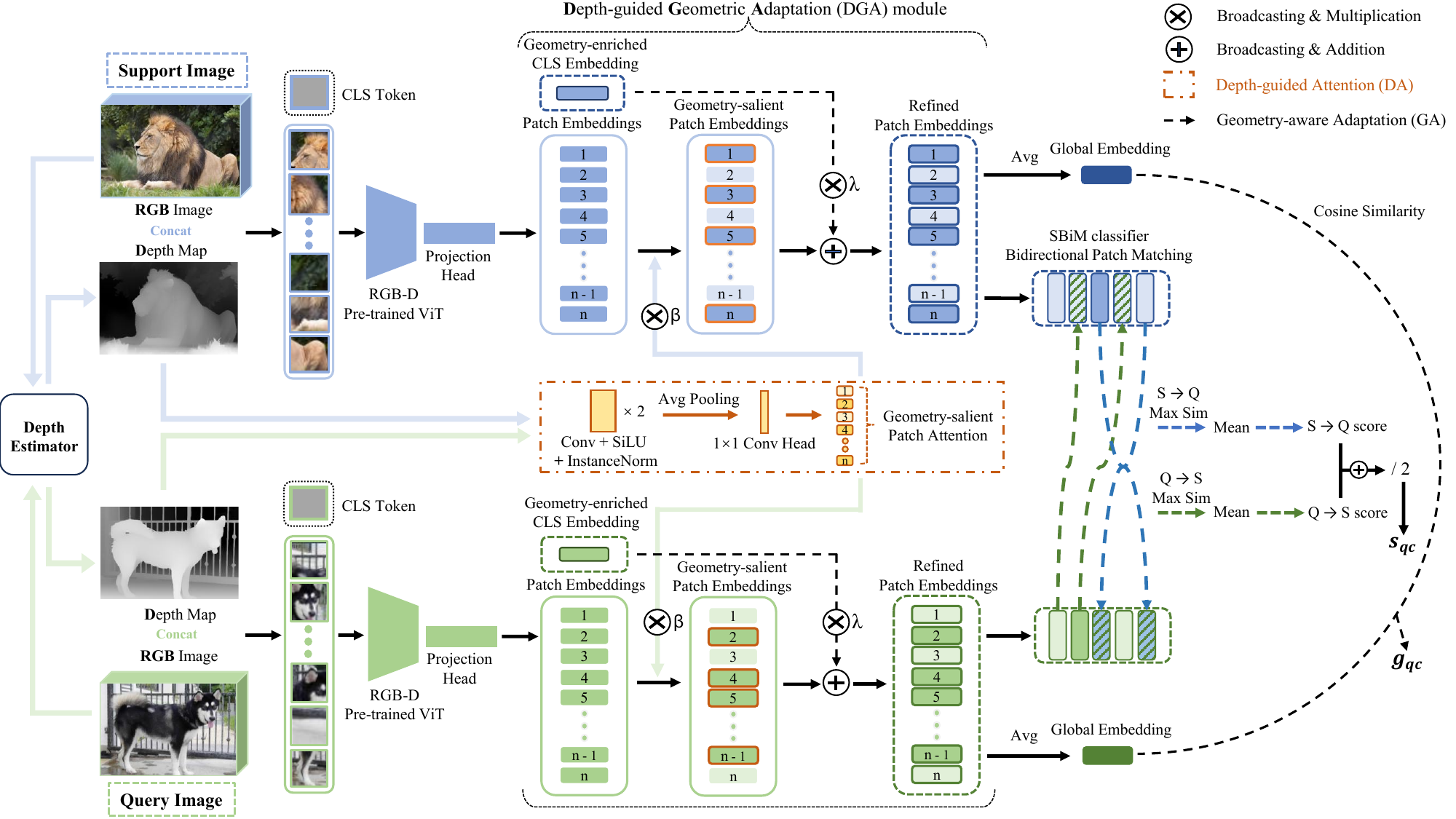}
    \caption{\textbf{Overview of our proposed DAFIC framework}. The depth estimator first predicts a pixel-wise aligned depth map, which is concatenated with the RGB image to form an RGB-D input. The RGB-D pre-trained ViT backbone then extracts token embeddings, where the CLS token encodes geometry-enriched global semantic context. Next, the \textit{Depth-guided Geometric Adaptation} (DGA) module refines the patch embeddings: \textit{Depth-guided Attention} (DA) converts depth priors into geometry-salient patch attention (scaled by $\beta$); \textit{Geometry-aware Adaptation} (GA) then integrates the enriched CLS (class) embedding (scaled by $\lambda$). Finally, based on the refined patch embeddings, the non-parametric SBiM classifier performs query-to-support bidirectional dense matching, together with an auxiliary global similarity objective, for reliable FSIC.}
    \label{fig:framework}
    \vspace{-5mm}
\end{figure*}

\section{Method}
We begin by formulating the FSIC problem. We then introduce DAFIC, which consists of three components: (1) an RGB-D self-supervised pre-training scheme for geometry-aware token embeddings; (2) a Depth-guided Geometric Adaptation (DGA) module that converts depth priors into geometry-salient patch attention and injects a geometry-enriched class embedding, enriching salient patches with class-level context while steering low-attention (often background) patches toward more consistent global descriptors; and (3) a Spatial Bidirectional Patch Matching (SBiM) classifier for non-parametric bidirectional dense matching. In addition, we introduce an auxiliary global supervision signal via spatial average pooling over the refined patch embeddings. During episodic training, the local and global objectives are jointly optimized for few-shot classification.

\paragraph{\textbf{Problem Formulation.}} Few-shot image classification aims to generalize well to novel classes with only a few labeled image samples. The FSIC setting commonly consists of two parts: a base set $\mathcal{D}_{base} = \{(I, y)|I\in \mathcal{X}_{base}, y\in \mathcal{C}_{base}\}$ for training and a novel set $\mathcal{D}_{novel} = \{(I, y)|I\in \mathcal{X}_{novel}, y\in \mathcal{C}_{novel}\}$ for few-shot testing, where $I$ denotes the image and $y$ represents the label. Notably, their label spaces are strictly disjoint, i.e., $\mathcal{C}_{base} \cap\mathcal{C}_{novel}=\emptyset$. During the meta-testing phase, a support set $\mathcal{S} = \{(I_i, y_i)\}_{i=1}^{N \times K}$ is randomly sampled from $\mathcal{D}_{novel}$, containing $N$ classes with $K$ labeled samples per class. The few-shot model is then required to correctly classify images from a query set $\mathcal{Q} = \{(I_i, y_i)\}_{i=1}^{N\times M}$ into one of the $N$ support classes, where $M$ is the number of query samples per class. This evaluation protocol is referred to as an $N$-way $K$-shot classification task.

\subsection{Self-supervised Pre-training with Depth Cues}
Inspired by recent advances in self-supervised Vision Transformer training~\cite{caron2021DINO,zhou2021ibot}, we adopt the teacher–student self-distillation pre-training framework with Masked Image Modeling (MIM)~\cite{beit,zhou2021ibot} as the pretext task. This paradigm effectively alleviates the risk of supervision collapse and has been shown to be beneficial in recent FSIC studies~\cite{hiller2022Fewtrue,hao2023CPEA,lin2023SMKD,cumplido2025slot}. Building on this, we extend this pre-training framework with depth-aided cues.

Distinct from standard RGB pre-training, we augment the input image $I_{\text{img}} \in \mathbb{R}^{H \times W \times 3}$ with a pixel-wise depth map $D_{\text{img}}$, estimated via a zero-shot monocular estimator $\mathcal{E}_{\text{depth}}$. The RGB and depth channels are individually normalized and concatenated to form the 4-channel input $I_{\text{rgbd}}$. To preserve geometric consistency, we implement a split augmentation strategy: spatial transformations (e.g., random crop, flip) are applied to both RGB and depth channels, whereas appearance transformations (e.g., color jitter) are applied exclusively to the RGB channels. This helps keep the structural depth cues consistent.

During pre-training, the student network is trained to align with the teacher’s class-token predictions to capture global semantic consistency ($\mathcal{L}_{\text{cls}}$), while simultaneously predicting masked patch representations by minimizing the divergence from the teacher’s predictions computed on unmasked tokens ($\mathcal{L}_{\text{mim}}$). Importantly, unlike standard MIM which can be biased toward appearance cues, predicting patch representations with depth information encourages the network to infer spatial layouts and object boundaries. Moreover, this structural understanding extends to the global level: via the self-attention mechanism, the [CLS] token integrates these geometric priors, thereby becoming a more geometry-aware global representation. The total objective aggregates this class-level self-distillation and the geometry-aware patch-level MIM loss:
\begin{equation}
\label{pretrain:total_loss}
\mathcal{L}_{\text{pretrain}} = \mathcal{L}_{\text{cls}} + \mathcal{L}_{\text{mim}},
\end{equation}
where both terms minimize the cross-entropy between the student's and teacher's predictions. Optimizing this joint objective encourages the model to capture occlusion- and layout-related cues, potentially reducing visual ambiguity in downstream few-shot tasks.

\subsection{Few-shot Learning with DGA Module}
As shown in Fig.~\ref{fig:framework}, given the input $I_{\text{rgbd}}$, the RGB-D pre-trained ViT backbone with projection head produces a $d$-dimensional token embedding sequence $\mathbf{X}\in \mathbb{R}^{(1+n)\times d}$:
\vspace{-1mm}
\begin{equation}
\mathbf{X} = [\mathbf{x}_0, \mathbf{x}_1, \dots, \mathbf{x}_{n}],
\label{eq:tokens}
\end{equation}
where $\mathbf{x}_0$ is the geometry-enriched class embedding and $\{\mathbf{x}_i\}_{i=1}^{n}$ denote the patch embeddings. However, raw patch embeddings often remain misaligned with global semantics and lack explicit guidance to separate target entities from background clutter. Therefore, we propose our depth-guided geometric adaptation module to address this via two sub-components: (1) depth-guided attention (DA) that transforms depth priors into geometry-salient patch attention, and (2) geometry-aware adaptation (GA) that integrates the geometry-enriched class embedding into the attention-modulated features to guide semantic alignment, thereby selectively enhancing the class relevance of geometry-salient patches for more reliable alignment.

\vspace{-3mm}
\paragraph{\textbf{Depth-guided attention.}}
First, the depth map $D_{\text{img}}$ is processed using min-max normalization to mitigate absolute scale variations across scenes while preserving relative geometric structure.
Then, the normalized depth map $\hat{D}_{\text{img}}$ is encoded by a lightweight two-layer convolutional neural network (CNN) module $\psi_{\text{dep}}$, which is optimized by episodic training. To accommodate the diverse depth distributions, we employ SiLU activations \cite{elfwing2018silu} and Instance Normalization \cite{ulyanov2016instancenorm} in each layer, followed by a $1\times1$ convolutional projection head. The resulting spatial feature map is then flattened to align with the sequence length of the ViT tokens, generating the geometry-salient patch attention scores $\mathbf{A}^{\text{dep}}\in \mathbb{R}^{n\times 1}$:
\vspace{-2mm}
\begin{equation}\label{eq:depth_attn}
\mathbf{A}^{\text{dep}} = \psi_{\text{dep}}\,\big(\hat{D}_{\text{img}}\big),
\end{equation}
which are then normalized via a patch-wise softmax operation to obtain the attention weights $\mathbf{W}^{\text{dep}}\in \mathbb{R}^{n\times 1}$:
\vspace{-0.5mm}
\begin{equation}\label{eq:depth_normalize}
\mathbf{W}^{\text{dep}}=\text{Softmax}(\mathbf{A}^{\text{dep}}).
\end{equation}

Next, each patch embedding $\mathbf{x}_i$ is modulated by its corresponding attention weight $W_i^{\text{dep}}$ and scaled by a learnable attention-strength parameter $\beta$ to obtain the geometry-salient patch embedding $\mathbf{x}_i^{\text{dep}}$:
\begin{equation}\label{eq:depth_modulate}
\mathbf{x}_i^{\text{dep}} \;=\; \beta \cdot W_i^{\text{dep}} \cdot \mathbf{x}_i,
\end{equation} 
where $i \in \{1, 2, \dots, n\}$. Here, $\beta$ acts as a gating mechanism to control the overall strength of the depth priors under the normalized softmax weighting. As a result, this modulation encourages emphasizing depth-inspired salient regions while suppressing the influence of background distractors.

\vspace{-3mm}
\paragraph{\textbf{Geometry-aware adaptation.}} To further align the local patches with the global semantics, we employ a global-to-local broadcasting mechanism. However, unlike prior methods \cite{hao2023CPEA,jiang2024CPES} that uniformly adapt all raw features, our method emphasizes geometry-salient patches via attention-modulated features from the preceding DA module. Moreover, unlike standard RGB models, our [CLS] token is derived from RGB-D MIM pre-training, serving as a geometry-enriched class embedding that encodes rich geometric cues. Consequently, this geometry-enriched global context $\mathbf{x}_0 \in \mathbb{R}^d$ is adapted to each geometry-salient patch embedding $\mathbf{x}_i^{\text{dep}}$ via:
\vspace{-2mm}
\begin{equation}\label{class_aware}
\tilde{\mathbf{x}}_i \;=\; \mathbf{x}_i^{\text{dep}} + \lambda\cdot \mathbf{x}_0,
\end{equation}
where $\lambda>0$ controls the magnitude of adaptation. Importantly, combined with the DA component, this operation improves local-to-global consistency: it encourages salient patches to preserve fine-grained local cues while injecting geometry-enriched class-level context, and steers low-attention (often background) patches toward a more consistent global representation dominated by $\mathbf{x}_0$. Finally, our DGA module produces refined patch embeddings $\tilde{\mathbf{X}}=[\tilde{\mathbf{x}}_1;\dots;\tilde{\mathbf{x}}_n]\in\mathbb{R}^{n\times d}$ for subsequent dense matching.

\begin{table*}[ht!]
    \centering
    \small
    \caption{Results (\%) on MiniImageNet and TieredImageNet. The ± shows 95\% confidence intervals. The best and second-best results are shown in \textbf{bold} and \underline{underlined}. A more comprehensive version of the table is shown in the supplementary materials.}
    \renewcommand{\arraystretch}{0.95}
    \begin{tabular}{c l c c c c c c}
        \toprule
         &\multirow{2}{*}{\textbf{Method}} & \multirow{2}{*}{\textbf{Backbone}} & \multirow{2}{*}{\textbf{$\approx$ Params}} 
         & \multicolumn{2}{c}{\textbf{MiniImageNet}} & \multicolumn{2}{c}{\textbf{TieredImageNet}}\\ 
        \cmidrule(lr){5-6}\cmidrule(lr){7-8}
         &  &  &  
         & \textbf{5-way 1-shot} & \textbf{5-way 5-shot}  
         & \textbf{5-way 1-shot} & \textbf{5-way 5-shot}\\
        \midrule
        &ProtoNet \cite{snell2017ProtoNet} & ResNet-12 & 12.4M & 62.29{\scriptsize ±0.33} & 79.46{\scriptsize ±0.48} & 68.25{\scriptsize ±0.23} & 84.01{\scriptsize ±0.56} \\
        &CAN \cite{hou2019cross} & ResNet-12 & 12.4M & 63.85{\scriptsize ±0.48} & 79.44{\scriptsize ±0.34} & 69.89{\scriptsize ±0.51} & 84.23{\scriptsize ±0.37} \\
        &FEAT \cite{ye2020FEAT} & ResNet-12 & 12.4M & 66.78{\scriptsize ±0.20} & 82.05{\scriptsize ±0.14} & 70.80{\scriptsize ±0.23} & 84.79{\scriptsize ±0.16}\\
        &RENet \cite{kang2021REnet} & ResNet-12 & 12.4M & 67.60{\scriptsize ±0.44} & 82.58{\scriptsize ±0.30} & 71.61{\scriptsize ±0.51} & 85.28{\scriptsize ±0.35} \\
        &SUN \cite{dong2022self} & ViT & 12.5M & 67.80{\scriptsize ±0.45} & 83.25{\scriptsize ±0.30} & 72.99{\scriptsize ±0.50} & 86.74{\scriptsize ±0.33}\\
        &FewTURE \cite{hiller2022Fewtrue} & Swin-T & 29M & 72.40{\scriptsize ±0.78} & 86.38{\scriptsize ±0.49} & 76.32{\scriptsize ±0.87} & 89.96{\scriptsize ±0.55} \\
        &CPEA \cite{hao2023CPEA} & ViT-S & 22M & 71.97{\scriptsize ±0.65} & 87.06{\scriptsize ±0.38} & 76.93{\scriptsize ±0.70} & 90.12{\scriptsize ±0.45} \\ 
        &SMKD \cite{lin2023SMKD} & ViT-S & 22M & \underline{74.28}{\scriptsize ±0.18} & \underline{88.82}{\scriptsize ±0.09} & \underline{78.83}{\scriptsize ±0.20} & \underline{91.02}{\scriptsize ±0.12} \\
        &SAFF \cite{cumplido2025slot} & ViT-S & 22M & 71.51{\scriptsize ±0.63} & 87.19{\scriptsize ±0.36} & 74.73{\scriptsize ±0.74} & 88.97{\scriptsize ±0.43} \\
        \cmidrule(lr){2-8}
        &DAFIC (Ours) & ViT-S & 22M & \textbf{77.71{\scriptsize ±0.44}} & \textbf{90.35{\scriptsize ±0.51}} & \textbf{80.10{\scriptsize ±0.40}} & \textbf{91.05{\scriptsize ±0.51}} \\
        \bottomrule
    \end{tabular}
    \label{tab:table1}
\end{table*}

\begin{table*}[ht!]
    \centering
    \small
    \caption{Results (\%) on CIFAR-FS and FC100.}
    \begin{tabular}{c l c c c c c c}
        \toprule
         &\multirow{2}{*}{\textbf{Method}} & \multirow{2}{*}{\textbf{Backbone}} & \multirow{2}{*}{\textbf{$\approx$ Params}} 
         & \multicolumn{2}{c}{\textbf{CIFAR-FS}} & \multicolumn{2}{c}{\textbf{FC100}}\\ 
        \cmidrule(lr){5-6}\cmidrule(lr){7-8}
         &  &  &  
         & \textbf{5-way 1-shot} & \textbf{5-way 5-shot}  
         & \textbf{5-way 1-shot} & \textbf{5-way 5-shot}\\
        \midrule
        
        &ProtoNet \cite{snell2017ProtoNet} & ResNet-12 & 12.4M & 72.20{\scriptsize ±0.70} & 83.50{\scriptsize ±0.50} & 41.54{\scriptsize ±0.76} & 57.08{\scriptsize ±0.76}\\ 
        &RFS \cite{tian2020rethinking} & ResNet-12 & 12.4M & 71.50{\scriptsize ±0.80} & 86.00{\scriptsize ±0.50} & 42.60{\scriptsize ±0.70} & 59.10{\scriptsize ±0.60} \\
        &SUN \cite{dong2022self} & ViT & 12.5M & 78.37{\scriptsize ±0.46} & 88.84{\scriptsize ±0.32} & - & -\\
        &FewTURE \cite{hiller2022Fewtrue} & Swin-T & 29M & 77.76{\scriptsize ±0.81} & 88.90{\scriptsize ±0.59} & 47.68{\scriptsize ±0.78} & 63.81{\scriptsize ±0.75} \\
        &CPEA \cite{hao2023CPEA} & ViT-S & 22M & 77.82{\scriptsize ±0.66} & 88.98{\scriptsize ±0.45} & 47.24{\scriptsize ±0.58} & 65.02{\scriptsize ±0.60}\\
        &SMKD \cite{lin2023SMKD} & ViT-S & 22M & \underline{80.08}{\scriptsize ±0.18} & \underline{90.63}{\scriptsize ±0.13} & \underline{50.38}{\scriptsize ±0.16} & \underline{68.37}{\scriptsize ±0.16} \\ 
        &SAFF \cite{cumplido2025slot} & ViT-S & 22M & 78.48{\scriptsize ±0.67} & 90.30{\scriptsize ±0.41} & 47.17{\scriptsize ±0.59} & 66.00{\scriptsize ±0.57} \\
        \cmidrule(lr){2-8}
        &DAFIC (Ours) & ViT-S & 22M & \textbf{84.30{\scriptsize ±0.44}} & \textbf{92.53{\scriptsize ±0.29}} & \textbf{50.52{\scriptsize ±0.46}} & \textbf{69.42{\scriptsize ±0.44}} \\
        \bottomrule
    \end{tabular}
    \vspace{-3mm}
    \label{tab:table2}
\end{table*}

\subsection{Few-shot Classification via Bidirectional Patch Matching}
\label{sec:classification}
The next challenge is to establish reliable dense comparisons between images. Unlike previous methods that flatten the spatial dimensions of features and map them through fixed parametric heads (e.g., linear layers or MLPs) for similarity measurement \cite{hao2023CPEA, jiang2024CPES, cumplido2025slot}, we adopt a non-parametric, symmetric patch-matching strategy, which retains patch-level correspondence without flattening patch-wise relations and achieves effective matching by leveraging the geometry-salient and class-aware patch embeddings $\tilde{\mathbf{X}}$.

Let $\tilde{\mathbf{X}}^{(Q)}_q \in \mathbb{R}^{n \times d}$ denote the refined feature representation of the $q$-th query image. For an $N$-way $K$-shot task, rather than naively averaging the support features spatially, which inevitably discards fine-grained local correspondence~\cite{hao2023CPEA, jiang2024CPES, cumplido2025slot}, we construct a dense support patch super-set $\tilde{\mathbf{X}}^{(S)}_c \in \mathbb{R}^{(Kn) \times d}$ by aggregating all $K \times n$ refined support patches belonging to class $c$. Then, we compute the dense patch-to-patch cosine similarity matrix $\mathbf{S}^{(q, c)} \in \mathbb{R}^{n \times (Kn)}$:
\vspace{-2mm}
\begin{equation}
    \label{eq:sim_matrix}
    \mathbf{S}^{(q, c)}_{i, j} = \frac{(\tilde{\mathbf{x}}^{(Q)}_{q, i})^\top \tilde{\mathbf{x}}^{(S)}_{c, j}}{\|\tilde{\mathbf{x}}^{(Q)}_{q, i}\|_2 \|\tilde{\mathbf{x}}^{(S)}_{c, j}\|_2},
\vspace{-1mm}
\end{equation}
where $\tilde{\mathbf{x}}^{(Q)}_{q, i}$ and $\tilde{\mathbf{x}}^{(S)}_{c, j}$ denote the $i$-th query patch and the $j$-th support patch in the support patch super-set, respectively. 

To achieve reliable local matching in the few-shot scenario, we adopt a Spatial Bidirectional Patch Matching (SBiM) classifier. Specifically, we measure the bidirectional correspondence by computing the Query-to-Support ($s_{q \to c}$) and Support-to-Query ($s_{c \to q}$) similarities. For the forward matching, each query patch seeks its nearest neighbor (maximum similarity) in the entire support patch super-set, followed by a spatial average over all $n$ query patches:
\vspace{-2mm}
\begin{equation}
    \label{eq:forward_match}
    s_{q \to c} = \frac{1}{n} \sum_{i=1}^{n} \max_{j \in \{1, \dots, Kn\}} \mathbf{S}^{(q, c)}_{i, j}.
\end{equation}
Symmetrically, backward matching ensures that every support patch finds its most relevant counterpart in the query image, averaged over the $Kn$ support patches:
\vspace{-2mm}
\begin{equation}
    \label{eq:backward_match}
    s_{c \to q} = \frac{1}{Kn} \sum_{j=1}^{Kn} \max_{i \in \{1, \dots, n\}} \mathbf{S}^{(q, c)}_{i, j}.
\end{equation}

The final robust relation score $s_{qc}$ between the $q$-th query and the $c$-th class is obtained by fusing the bidirectional matching scores:
\vspace{-3mm}
\begin{equation}
    \label{eq:final_score}
    {s}_{qc} = \frac{s_{q \to c} + s_{c \to q}}{2}.
\end{equation}
By symmetrizing the matching scores (Eqs.~\ref{eq:forward_match} and \ref{eq:backward_match}), SBiM captures complementary patch-coverage information and helps mitigate spurious correspondences arising from one-way matching. More importantly, building on the frontend DGA module that highlights salient, class-aware entities and steers distractors toward more consistent global descriptors via geometric priors, SBiM further alleviates many-to-one matching issues. With Eq.~\ref{class_aware}, low-attention patches (small $\beta W_i^{\text{dep}}$) yield embeddings dominated by the global component $\lambda\mathbf{x}_0$. Since cosine similarity is scale-invariant, dense matching becomes less sensitive to feature magnitudes, making these contributions more consistent and less prone to clutter-induced spurious matches.

We further leverage the refined representations $\tilde{\mathbf{X}}$ to introduce an auxiliary global loss, providing holistic image-level supervision. Specifically, we apply spatial average pooling over $\tilde{\mathbf{X}}$ to derive $\ell_2$-normalized global embeddings $\bar{\mathbf{x}}^{(Q)}_q \in \mathbb{R}^d$ for the queries and global prototypes $\bar{\mathbf{x}}^{(S)}_c \in \mathbb{R}^d$ for the support classes. Then a global similarity score $g_{qc}$ is calculated as:
\vspace{-2mm}
\begin{equation}
    \label{eq:global_sim}
    g_{qc} = \frac{(\bar{\mathbf{x}}^{(Q)}_q)^\top \bar{\mathbf{x}}^{(S)}_c}{\|\bar{\mathbf{x}}^{(Q)}_q\|_2 \|\bar{\mathbf{x}}^{(S)}_c\|_2}.
\end{equation}
Finally, all trainable DAFIC components are jointly optimized via episodic training. The overall objective jointly minimizes the local cross-entropy loss $\mathcal{L}_{\text{local}}$ (derived from ${s}_{qc}$) and the global cross-entropy loss $\mathcal{L}_{\text{global}}$ (derived from $g_{qc}$):
\begin{equation}
    \label{eq:FSL}
    \mathcal{L}_{\text{FSL}} = \alpha\mathcal{L}_{\text{local}} + \gamma\mathcal{L}_{\text{global}},
\end{equation}
where $\alpha$ and $\gamma$ control the contribution of the local and global objectives. During inference, we use ${s}_{qc}$ for prediction, and $g_{qc}$ serves as an auxiliary training signal.

\section{Experiments}

We first detail the experimental settings and then compare our method with state-of-the-art approaches. Finally, we ablate the key components of our DAFIC framework to validate its effectiveness in the FSIC scenario.

\subsection{Experimental Settings}
\textbf{Datasets.} We conduct extensive experiments on four popular FSIC benchmarks: MiniImageNet~\cite{vinyals2016MatchingNet}, TieredImageNet~\cite{ren2018tieredImageNet}, CIFAR-FS~\cite{bertinetto2018CifarFS}, and FC100~\cite{oreshkin2018FC100}. We strictly follow the class splits of previous works \cite{hiller2022Fewtrue,hao2023CPEA, lin2023SMKD, jiang2024CPES, cumplido2025slot} for training, validation, and testing: 64/16/20 for both MiniImageNet and CIFAR-FS, 351/97/160 for TieredImageNet, and 60/20/20 for FC100.

\noindent\textbf{Depth Map Generation.} We use the DepthAnything V2-base~\cite{yang2024depth2} as an off-the-shelf estimator to generate monocular depth maps via a single feed-forward inference pass. Since the benchmark image set is fixed, we precompute and cache depth maps for all images to avoid redundant computation during episodic training and evaluation. These depth maps serve as a task-agnostic geometric prior for DAFIC.

\noindent\textbf{Backbone and Pre-training Setup.} Following prior FSIC works~\cite{hiller2022Fewtrue,hao2023CPEA,cumplido2025slot}, we adopt ViT-S/16~\cite{dosovitskiy2020ViT} as the backbone,with $224\times224$ inputs and 384-dimensional token embeddings. Our RGB-D pre-training generally converges faster and improves downstream few-shot classification performance over RGB-only pre-training~\cite{hiller2022Fewtrue,hao2023CPEA}. For pre-training, we use three NVIDIA A100-40GB GPUs, batch size 384, and 1,000 epochs, compared with four GPUs, batch size 512, and 1,600 epochs used by RGB-only counterparts~\cite{hiller2022Fewtrue,hao2023CPEA}. Detailed configurations and pre-training comparisons are provided in the supplementary materials.

\noindent\textbf{Training and Implementation Details.} Following the learning paradigm of most recent FSIC works~\cite{ye2020FEAT,hiller2022Fewtrue,hao2023CPEA,lin2023SMKD,cumplido2025slot}, our training consists of two stages: we first pre-train the ViT-S/16 backbone from scratch using $\mathcal{L}_{\text{pretrain}}$ in Eq.~\eqref{pretrain:total_loss}, and then fine-tune the entire framework with episodic learning using $\mathcal{L}_{\text{FSL}}$ in Eq.~\eqref{eq:FSL}. Importantly, all learnable components of DAFIC in both stages are optimized exclusively on the meta-train split. For the second-stage FSL optimization, we adopt the Adam optimizer~\cite{diederik2014adam} with an initial learning rate of $0.001$, which is halved every $1,000$ episodes. The backbone is updated with a learning rate of $1 \times 10^{-5}$ and a weight decay of $0.01$ over $10{,}000$ training episodes. We set $\lambda=2.0$, initialize the learnable $\beta$ to $5.0$ to provide sufficient early gate strength, and empirically set the fixed weights $\alpha=1.0$ and $\gamma=0.5$.

\noindent\textbf{Evaluation Protocol.} The performance evaluation follows the standard $5$-way $1$-shot and $5$-shot FSIC protocols. We report mean accuracy across 1,000 random episodes (15 queries per class).

\begin{figure}
    \centering
    \includegraphics[width=1.0\columnwidth, trim=22mm 5mm 20mm 3mm, clip]{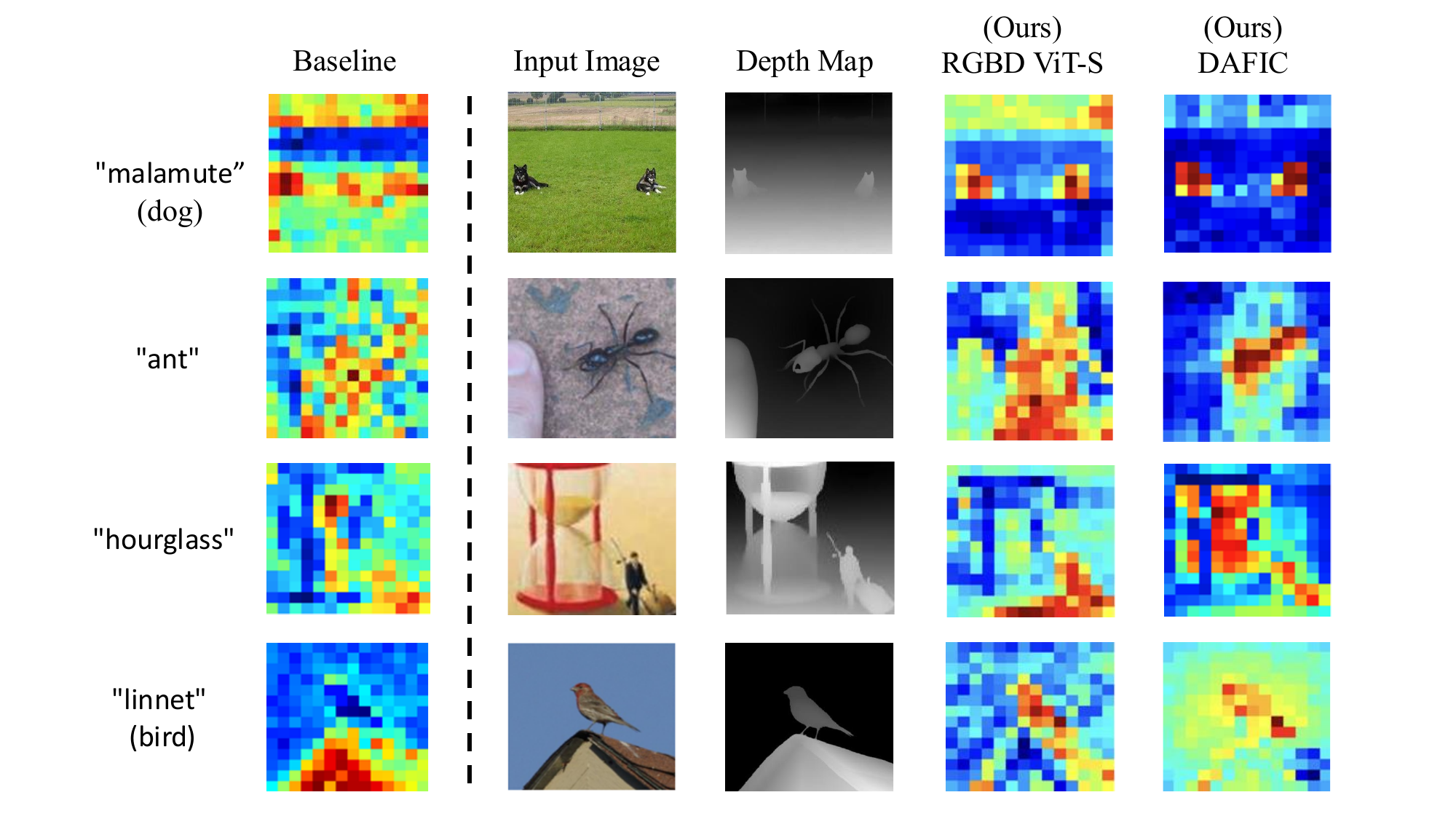}
    \caption{Comparison of patch-to-class semantic alignment between baseline and DAFIC. The "Baseline" and "DAFIC" columns visualize the alignment after adaptation, where red indicates higher class relevance and blue indicates lower relevance.}
    \label{fig:comparison_patch_attention}
\vspace{-2mm}
\end{figure}

\begin{table}[t]
\centering
\scriptsize
\caption{Controlled comparison of the reproduced CPEA baseline and DAFIC on four FSIC benchmarks. CPEA(DGA) and CPEA(DGA+SBiM) progressively incorporate DGA and SBiM into the baseline. Results are mean accuracy (\%) over 1,000 meta-test episodes; $\pm$ denotes 95\% confidence intervals.}
\label{tab:comparison_mean}
\setlength{\tabcolsep}{2pt} 
\renewcommand{\arraystretch}{0.95} 
\resizebox{0.99\columnwidth}{!}{
\begin{tabular}{lccccc}
\toprule

\textbf{Dataset} & \textbf{Shot} & \textbf{CPEA} & \textbf{CPEA(DGA)} & \textbf{ CPEA(DGA+SBiM)} & \textbf{DAFIC} \\
\midrule
\multirow{2}{*}{MiniImageNet\cite{vinyals2016MatchingNet}} 
& 1 & 72.25{\scriptsize ±0.65} & 73.26{\scriptsize ±0.60} & 74.02{\scriptsize ±0.46} & \textbf{77.71}{\scriptsize ±0.44} \\
& 5 & 87.24{\scriptsize ±0.38} & 87.82{\scriptsize ±0.33} & 88.12{\scriptsize ±0.27} & \textbf{90.35}{\scriptsize ±0.51} \\
\midrule
\multirow{2}{*}{TieredImageNet\cite{ren2018tieredImageNet}} 
& 1 & 77.12{\scriptsize ±0.70} & 78.30{\scriptsize ±0.70} & 78.86{\scriptsize ±0.51} & \textbf{80.10}{\scriptsize ±0.40} \\
& 5 & 90.12{\scriptsize ±0.45} & 90.65{\scriptsize ±0.41} & 90.61{\scriptsize ±0.32} & \textbf{91.05}{\scriptsize ±0.51} \\
\midrule
\multirow{2}{*}{CIFAR-FS\cite{bertinetto2018CifarFS}} 
& 1 & 78.27{\scriptsize ±0.66} & 78.66{\scriptsize ±0.60} & 79.05{\scriptsize ±0.47} & \textbf{84.30}{\scriptsize ±0.44} \\
& 5 & 89.30{\scriptsize ±0.45} & 89.92{\scriptsize ±0.41} & 89.99{\scriptsize ±0.33} & \textbf{92.53}{\scriptsize ±0.29} \\
\midrule
\multirow{2}{*}{FC100\cite{oreshkin2018FC100}} 
& 1 & 46.51{\scriptsize ±0.58} & 47.62{\scriptsize ±0.60} & 49.31{\scriptsize ±0.46} & \textbf{50.52}{\scriptsize ±0.46} \\
& 5 & 65.18{\scriptsize ±0.60} & 66.54{\scriptsize ±0.59} & 67.07{\scriptsize ±0.42} & \textbf{69.42}{\scriptsize ±0.44} \\
\bottomrule
\end{tabular}
}
\vspace{-3mm}
\end{table}

{\subsection{Comparison results}

Tab.~\ref{tab:table1} and Tab.~\ref{tab:table2} compare DAFIC with recent state-of-the-art methods under 5-way 1-shot and 5-shot settings on four benchmark datasets. DAFIC achieves strong performance, especially in the data-scarce 1-shot setting. Compared with the CPEA baseline~\cite{hao2023CPEA}, DAFIC improves accuracy by +5.74\% on MiniImageNet (77.71\% vs. 71.97\%) and +6.48\% on CIFAR-FS (84.30\% vs. 77.82\%), showing the benefit of introducing geometric priors beyond RGB-only representations. DAFIC also outperforms the strongest recent competitor, SMKD~\cite{lin2023SMKD}, by +3.43\%/+4.22\% in the 1-shot setting and +1.53\%/+1.90\% in the 5-shot setting on MiniImageNet/CIFAR-FS.

Qualitatively, Fig.~\ref{fig:comparison_patch_attention} visualizes the patch-to-class semantic alignment of our method against the baseline~\cite{hao2023CPEA}. More visualizations are provided in the supplementary materials. DAFIC demonstrates advantages in localizing class-relevant regions while suppressing background noise. Notably, even without any few-shot adaptation, the raw patch embeddings extracted directly from our RGB-D pre-trained backbone exhibit a stronger class-aware focus than the baseline. Building upon this strong initialization, the complete DAFIC framework further refines these representations to align with target object regions. Overall, both quantitative and qualitative results demonstrate that the design of DAFIC mitigates visual ambiguity in FSIC scenarios.

\subsection{Ablation Study}
We assess the contributions of DAFIC components and the impact of monocular depth priors. Additional ablations are provided in Section D of the supplementary material.

\noindent\textbf{Effect of depth on representation learning.} As shown in Fig. \ref{fig:Sup_UnSup_UnSupD}, our RGB-D pre-training obtains gains of +3.25\% and +2.48\% on 1-shot and 5-shot tasks over RGB-only counterparts~\cite{hiller2022Fewtrue, hao2023CPEA}, even when using a simple prototypical classifier. This demonstrates the effectiveness of integrating depth into self-supervised MIM to enrich the representation space and provide a strong initialization for the following few-shot learning.

\begin{figure}[t]
        \centering
        \includegraphics[width=.60\linewidth]{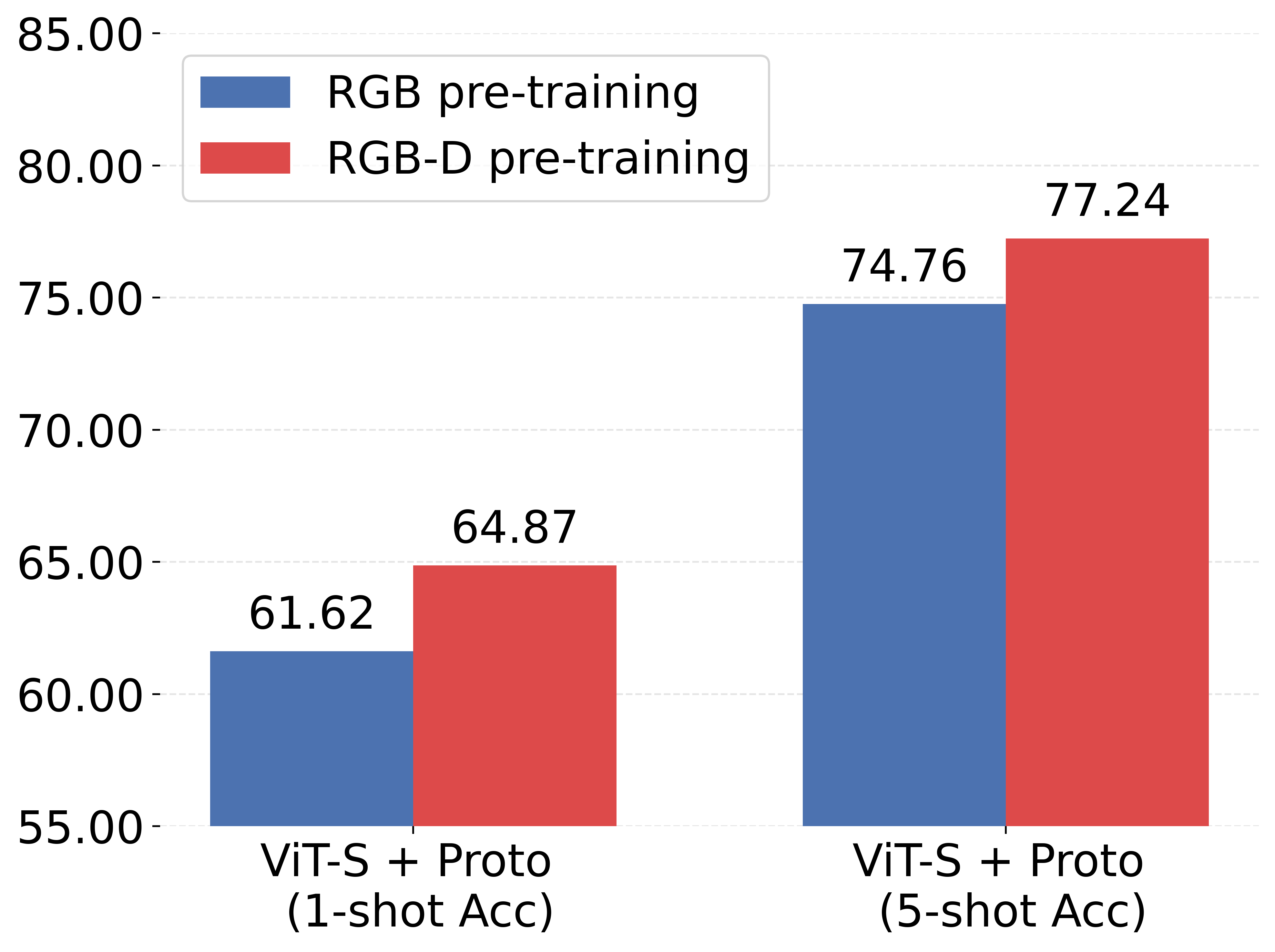}
        \caption{Comparison of RGB- and RGB-D-based pre-training on MiniImageNet. The bars show average test accuracies (\%) under 5-way 1/5-shot settings, with a simple prototypical classifier.}
        \vspace{-2mm}
        \label{fig:Sup_UnSup_UnSupD}
\end{figure}

\begin{table}[t]
\centering
\scriptsize
\setlength{\tabcolsep}{4pt}
\renewcommand{\arraystretch}{0.95}
\caption{DAFIC components ablation on MiniImageNet (5-way) with mean accuracy (\%) averaged over 1,000 meta-test episodes.}
\label{tab:DAFIC_components}

\resizebox{0.60\columnwidth}{!}{%
    \begin{tabular}{lcc}
        \toprule
        \textbf{Components} & \textbf{1-shot} & \textbf{5-shot} \\
        \midrule
        Non-pretrain  & 30.49{\scriptsize $\pm$0.35} & 41.23{\scriptsize $\pm$0.38} \\
        RGB-pretrain  & 69.12{\scriptsize $\pm$0.49} & 84.88{\scriptsize $\pm$0.31} \\
        RGBD-pretrain & 73.85{\scriptsize $\pm$0.48} & 87.64{\scriptsize $\pm$0.29} \\
        + DA          & 74.64{\scriptsize $\pm$0.47} & 88.38{\scriptsize $\pm$0.28} \\
        + GA          & 76.69{\scriptsize $\pm$0.51} & 89.45{\scriptsize $\pm$0.35} \\
        + SBiM        & 77.42{\scriptsize $\pm$0.49} & 90.14{\scriptsize $\pm$0.41} \\
        + Global Sup  & 77.71{\scriptsize $\pm$0.44} & 90.35{\scriptsize $\pm$0.51} \\
        \bottomrule
    \end{tabular}%
}
\vspace{-3mm}
\end{table}

\noindent\textbf{Effectiveness of DGA and SBiM.} Tab.~\ref{tab:comparison_mean} demonstrates the performance gains from integrating our proposed components. Incorporating CPEA with our DGA module yields consistent improvements across datasets (e.g., +1.18\% on 1-shot TieredImageNet; +1.36\% on 5-shot FC100). Further incorporating SBiM generally improves performance (+1.69\% on 1-shot FC100; +0.76\% on 1-shot MiniImageNet). Finally, our complete DAFIC framework with RGB-D initialization achieves the best performance, outperforming the DGA+SBiM variant (+3.69\% on MiniImageNet; +5.25\% on CIFAR-FS for 1-shot). Overall, these results across benchmarks support the effectiveness and complementarity of DGA, SBiM, and the overall framework.

\begin{table}[t] 
    \centering
    \caption{Computational complexity and performance comparison. FLOPs and inference latency are reported exclusively for the few-shot model, with latency averaged over 1800 query images on a single NVIDIA V100 GPU. Parameters (Params) correspond to the ViT-Small backbone. Performance is evaluated on CIFAR-FS.}
    \vspace{-2mm}
    \resizebox{0.99\linewidth}{!}{
        \begin{tabular}{@{} l c c c c c @{}} 
        \toprule
        \textbf{Model} & \textbf{Params (M)} & \textbf{FLOPs (G)} & \textbf{Time (ms)} & \textbf{1-shot Acc.}  & \textbf{5-shot Acc.}\\
        \midrule
        SMKD~\cite{lin2023SMKD} & 21.67 & 339.87 & 78.40 & 80.08±0.18 & 90.63±0.13 \\
        CPEA~\cite{hao2023CPEA} & 21.67 & 9.38 & 1.43 & 78.27±0.66 & 89.30±0.45\\
        DAFIC & 21.76 & 9.51 & 1.39 & 84.30±0.44 &  92.53±0.29 \\
        \bottomrule
        \end{tabular}
    }
    \label{tab:inference}
    \vspace{-4mm}
\end{table}

\noindent\textbf{Ablation of DAFIC components.} Tab.~\ref{tab:DAFIC_components} ablates DAFIC on MiniImageNet. Rows 1-5 use the same MLP dense matching classifier, with our components progressively added from Row 4. Without pre-training, the model performs poorly, highlighting the importance of ViT initialization in FSIC scenarios. Standard RGB pre-training improves the 1-shot accuracy to 69.12\%, while RGB-D pre-training further raises it to 73.85\% (+4.73\%), confirming the value of depth cues in enriching feature representations. Building upon the RGB-D representation, we observe steady improvements as each module is sequentially introduced. Integrating the DA component improves results by +0.79\% by refining local depth-guided focus, and subsequently adding the GA component further enhances performance by adapting global geometry-enriched context, yielding a clear +2.05\% improvement (reaching 76.69\%). Finally, incorporating SBiM and auxiliary Global Supervision (Global Sup) adds +0.73\% and +0.29\% respectively, completing our DAFIC architecture. Overall, the gain of DAFIC does not come solely from RGB-D pre-training; the subsequent DA, GA, SBiM, and Global Supervision modules further improve performance by +3.86\%, demonstrating that the proposed adaptation and matching components effectively transform depth-aware representations into more discriminative support-query matching.

\noindent\textbf{Inference latency and FLOPs analysis.}
In Tab.~\ref{tab:inference}, we evaluate the computational efficiency of DAFIC alongside its performance gains. DAFIC outperforms the baseline by 6.03\% (1-shot) and 3.23\% (5-shot) on CIFAR-FS, while maintaining comparable inference latency with negligible additional overhead under the cached-depth setting. Compared with SMKD, DAFIC also requires fewer FLOPs and achieves lower inference latency. As specified in the experimental setting, depth maps are precomputed offline and cached for episodic inference; thus, the latency reported in Tab.~\ref{tab:inference} corresponds to the cached-depth setting. We additionally report the full inference latency under both cached and on-the-fly depth estimation settings in Supp.Tab.18.

\noindent\textbf{Robustness to Depth Corruptions.} We acknowledge that monocular depth estimation can be unreliable in challenging visual conditions. In Fig.~\ref{fig:depth_manu_fig} and Tab.~\ref{tab:depth_manu_table}, we perform depth corruption experiments on MiniImageNet to simulate missing or unreliable geometric cues. Background masking incurs only a minor performance change (e.g., a $0.61\%$ drop in 5-way 1-shot), suggesting that our depth-guided module is not overly sensitive to background depth. More importantly, under object-region occlusion or low-quality depth maps, DAFIC exhibits graceful degradation and still outperforms the CPEA baseline. For example, with object-region occlusion, DAFIC retains $73.06\%$ 1-shot accuracy, maintaining a $+0.81\%$ margin over CPEA ($72.25\%$). With low-quality depth, it still achieves $74.53\%$, surpassing CPEA by $+2.28\%$. Moreover, Supp.Tab.19 further evaluates DAFIC using the weaker depth prior consistently throughout training and evaluation, while still maintaining effective few-shot performance. Overall, these results suggest that DAFIC remains relatively robust to degraded geometric priors across the evaluated settings.

\begin{table}[t] 
    \centering
    \includegraphics[width=0.99\columnwidth, trim=15mm 100mm 45mm 0mm, clip]{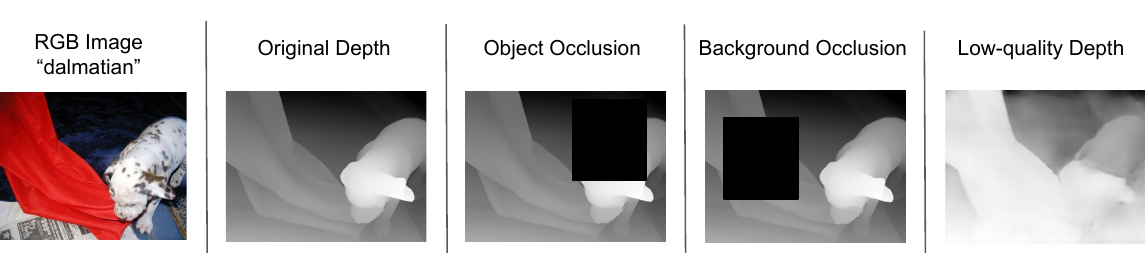}
    \makeatletter\def\@captype{figure}\makeatother 
    \caption{Example visualization of depth corruption strategies used during inference.}
    \label{fig:depth_manu_fig}
    \vspace{-4mm}
\end{table}

\begin{table}[t] 
    \centering
    \makeatletter\def\@captype{table}\makeatother 
    \caption{Inference robustness to corrupted depth on MiniImageNet. The CPEA results are reproduced.}
    \scriptsize
    \label{tab:depth_manu_table}
    \setlength{\tabcolsep}{6pt}
    \resizebox{0.90\linewidth}{!}{
        \begin{tabular}{l c c}
            \toprule
            \textbf{Methods} & \textbf{5-way 1-shot} & \textbf{5-way 5-shot} \\
            \midrule
            CPEA & 72.25±0.65 & 87.24±0.38 \\
            DAFIC & 77.71±0.44 & 90.35±0.51 \\
            DAFIC $\Rightarrow$ Object-region occlusions & 73.06±0.61 & 87.51±0.43 \\
            DAFIC $\Rightarrow$ Background occlusions & 77.10±0.64 & 89.81±0.36 \\
            DAFIC $\Rightarrow$ Low-quality depth map & 74.53±0.65 & 88.10±0.43 \\
            \bottomrule
        \end{tabular}
    }
\vspace{-3mm}
\end{table}

\section{Conclusion}
In this paper, we propose DAFIC, a geometry-guided FSIC framework that integrates monocular depth into representation learning, episodic adaptation, and dense support-query matching to mitigate \emph{appearance ambiguity} and \emph{background clutter}. DAFIC combines RGB-D self-supervised pre-training with Depth-guided Geometric Adaptation (DGA), which refines geometry-salient patches with geometry-enriched class context while suppressing background interference. Spatial Bidirectional Patch Matching (SBiM) then performs bidirectional support-query matching on the refined features to reduce spurious correspondences. Extensive experiments on four benchmarks show that DAFIC achieves strong performance against existing methods, highlighting the value of coupling geometry-conditioned episodic adaptation with bidirectional matching for few-shot visual recognition.

{
    \small
    \bibliographystyle{ieeenat_fullname}
    \bibliography{main}

@String(IJCV = {Int. J. Comput. Vis.})

@String(CVPR= {IEEE Conf. Comput. Vis. Pattern Recog.})

@String(ICCV= {Int. Conf. Comput. Vis.})

@String(ECCV= {Eur. Conf. Comput. Vis.})

@String(BMVC= {Brit. Mach. Vis. Conf.})

@String(ICLR = {Int. Conf. Learn. Represent.})

@String(AAAI = {AAAI})

@String(CVPRW= {IEEE Conf. Comput. Vis. Pattern Recog. Worksh.})

@String(IJCV  = {IJCV})

@String(CVPR  = {CVPR})

@String(ICCV  = {ICCV})

@String(ECCV  = {ECCV})

@String(BMVC  =	{BMVC})

@String(ICLR  = {ICLR})

@String(CVPRW= {CVPRW})

@inproceedings{wang2022multimodal,
  title={Multimodal token fusion for vision transformers},
  author={Wang, Yikai and Chen, Xinghao and Cao, Lele and Huang, Wenbing and Sun, Fuchun and Wang, Yunhe},
  booktitle={Proceedings of the IEEE/CVF Conference on Computer Vision and Pattern Recognition (CVPR)},
  pages={12186--12195},
  year={2022}
}

@article{zhang2023cmx,
  title={CMX: Cross-modal fusion for RGB-X semantic segmentation with transformers},
  author={Zhang, Jiaming and Liu, Huayao and Yang, Kailun and Hu, Xinxin and Liu, Ruiping and Stiefelhagen, Rainer},
  journal={IEEE Transactions on Intelligent Transportation Systems},
  volume={24},
  number={12},
  pages={14679--14694},
  year={2023},
  publisher={IEEE}
}

@inproceedings{yin2024dformer,
  title={DFormer: Rethinking RGBD Representation Learning for Semantic Segmentation},
  author={Yin, Bowen and Zhang, Xuying and Li, Zhong-Yu and Liu, Li and Cheng, Ming-Ming and Hou, Qibin},
  booktitle={International Conference on Learning Representations (ICLR)},
  year={2024}
}

@article{li2023libfewshot,
  title={Libfewshot: A comprehensive library for few-shot learning},
  author={Li, Wenbin and Wang, Ziyi and Yang, Xuesong and Dong, Chuanqi and Tian, Pinzhuo and Qin, Tiexin and Huo, Jing and Shi, Yinghuan and Wang, Lei and Gao, Yang and others},
  journal={IEEE Transactions on Pattern Analysis and Machine Intelligence},
  volume={45},
  number={12},
  pages={14938--14955},
  year={2023},
  publisher={IEEE}
}

@article{russakovsky2015imagenet,
  title={Imagenet large scale visual recognition challenge},
  author={Russakovsky, Olga and Deng, Jia and Su, Hao and Krause, Jonathan and Satheesh, Sanjeev and Ma, Sean and Huang, Zhiheng and Karpathy, Andrej and Khosla, Aditya and Bernstein, Michael and others},
  journal={International Journal of Computer Vision (IJCV)},
  volume={115},
  number={3},
  pages={211--252},
  year={2015},
  publisher={Springer}
}

@article{wang2020generalizing,
  title={{Generalizing from a few examples: A survey on few-shot learning}},
  author={Wang, Yaqing and Yao, Quanming and Kwok, James T and Ni, Lionel M},
  journal={ACM Computing Surveys},
  volume={53},
  number={3},
  pages={1--34},
  year={2020},
  publisher={ACM New York, NY, USA}
}

@article{doersch2020CTX,
  title={Crosstransformers: spatially-aware few-shot transfer},
  author={Doersch, Carl and Gupta, Ankush and Zisserman, Andrew},
  journal={Advances in Neural Information Processing Systems},
  volume={33},
  pages={21981--21993},
  year={2020}
}

@inproceedings{finn2017model,
  title={Model-agnostic meta-learning for fast adaptation of deep networks},
  author={Finn, Chelsea and Abbeel, Pieter and Levine, Sergey},
  booktitle={Proceedings of International Conference on Machine Learning (ICML)},
  pages={1126--1135},
  year={2017}
}

@article{nichol2018first,
  title={On first-order meta-learning algorithms},
  author={Nichol, Alex and Achiam, Joshua and Schulman, John},
  journal={arXiv preprint arXiv:1803.02999},
  year={2018}
}

@inproceedings{zintgraf2019fast,
  title={Fast context adaptation via meta-learning},
  author={Zintgraf, Luisa and Shiarli, Kyriacos and Kurin, Vitaly and Hofmann, Katja and Whiteson, Shimon},
  booktitle={Proceedings of International Conference on Machine Learning (ICML)},
  pages={7693--7702},
  year={2019}
}

@inproceedings{
rusu2018meta,
title={Meta-Learning with Latent Embedding Optimization},
author={Andrei A. Rusu and Dushyant Rao and Jakub Sygnowski and Oriol Vinyals and Razvan Pascanu and Simon Osindero and Raia Hadsell},
booktitle={International Conference on Learning Representations (ICLR)},
year={2019}}

@inproceedings{chen2019closer,
   author = {Chen, Wei-Yu and Liu, Yen-Cheng and Kira, Zsolt and Wang, Yu-Chiang and  Huang, Jia-Bin},
   title = {A Closer Look at Few-shot Classification},
   booktitle = {International Conference on Learning Representations (ICLR)},
   year = {2019}
}

@inproceedings{rajasegaran2020self,
  title={Self-supervised knowledge distillation for few-shot learning},
  author={Rajasegaran, Jathushan and Khan, Salman and Hayat, Munawar and Khan, Fahad Shahbaz and Shah, Mubarak},
  booktitle={Proceedings of British Machine Vision Conference (BMVC)},
  year={2021}
}

@inproceedings{mangla2020charting,
  title={Charting the right manifold: Manifold mixup for few-shot learning},
  author={Mangla, Puneet and Kumari, Nupur and Sinha, Abhishek and Singh, Mayank and Krishnamurthy, Balaji and Balasubramanian, Vineeth N},
  booktitle={Proceedings of the IEEE Winter Conference on Applications of Computer Vision (WACV)},
  pages={2218--2227},
  year={2020}
}

@inproceedings{liu2020negative,
  title={{Negative margin matters: Understanding margin in few-shot classification}},
  author={Liu, Bin and Cao, Yue and Lin, Yutong and Li, Qi and Zhang, Zheng and Long, Mingsheng and Hu, Han},
  booktitle={European Conference on Computer Vision (ECCV)},
  pages={438--455},
  year={2020},
  organization={Springer}
}

@article{vinyals2016MatchingNet,
  title={Matching networks for one-shot learning},
  author={Vinyals, Oriol and Blundell, Charles and Lillicrap, Timothy and Kavukcuoglu, Koray and Wierstra, Daan},
  journal={Advances in Neural Information Processing Systems},
  volume={29},
  year={2016}
}

@article{snell2017ProtoNet,
  title={Prototypical networks for few-shot learning},
  author={Snell, Jake and Swersky, Kevin and Zemel, Richard},
  journal={Advances in Neural Information Processing Systems},
  volume={30},
  year={2017}
}

@article{zhang2022DeepEMD,
  title={{Deepemd: Differentiable Earth mover's distance for few-shot learning}},
  author={Zhang, Chi and Cai, Yujun and Lin, Guosheng and Shen, Chunhua},
  journal={IEEE Transactions on Pattern Analysis and Machine Intelligence},
  volume={45},
  number={5},
  pages={5632--5648},
  year={2023},
  publisher={IEEE}
}

@inproceedings{Li2019DN4,
  title={Revisiting local descriptor based image-to-class measure for few-shot learning},
  author={Li, Wenbin and Wang, Lei and Xu, Jinglin and Huo, Jing and Gao, Yang and Luo, Jiebo},
  booktitle={Proceedings of the IEEE/CVF Conference on Computer Vision and Pattern Recognition (CVPR)},
  pages={7260--7268},
  year={2019}
}

@inproceedings{li2019CTM,
  title={Finding task-relevant features for few-shot learning by category traversal},
  author={Li, Hongyang and Eigen, David and Dodge, Samuel and Zeiler, Matthew and Wang, Xiaogang},
  booktitle={Proceedings of the IEEE/CVF Conference on Computer Vision and Pattern Recognition (CVPR)},
  pages={1--10},
  year={2019}
}

@inproceedings{ye2020FEAT,
  title={Few-shot learning via embedding adaptation with set-to-set functions},
  author={Ye, Han-Jia and Hu, Hexiang and Zhan, De-Chuan and Sha, Fei},
  booktitle={Proceedings of the IEEE/CVF Conference on Computer Vision and Pattern Recognition (CVPR)},
  pages={8808--8817},
  year={2020}
}

@inproceedings{wertheimer2021FRN,
  title={Few-shot classification with feature map reconstruction networks},
  author={Wertheimer, Davis and Tang, Luming and Hariharan, Bharath},
  booktitle={Proceedings of the IEEE/CVF Conference on Computer Vision and Pattern Recognition (CVPR)},
  pages={8012--8021},
  year={2021}
}

@inproceedings{cheng2023FGFL,
  title={Frequency guidance matters in few-shot learning},
  author={Cheng, Hao and Yang, Siyuan and Zhou, Joey Tianyi and Guo, Lanqing and Wen, Bihan},
  booktitle={Proceedings of the IEEE/CVF International Conference on Computer Vision (ICCV)},
  pages={11814--11824},
  year={2023}
}

@article{hiller2022Fewtrue,
  title={Rethinking generalization in few-shot classification},
  author={Hiller, Markus and Ma, Rongkai and Harandi, Mehrtash and Drummond, Tom},
  journal={Advances in Neural Information Processing Systems},
  volume={35},
  pages={3582--3595},
  year={2022}
}

@inproceedings{hao2023CPEA,
  title={Class-aware patch embedding adaptation for few-shot image classification},
  author={Hao, Fusheng and He, Fengxiang and Liu, Liu and Wu, Fuxiang and Tao, Dacheng and Cheng, Jun},
  booktitle={Proceedings of the IEEE/CVF International Conference on Computer Vision (ICCV)},
  pages={18905--18915},
  year={2023}
}

@inproceedings{dong2022self,
  title={Self-promoted supervision for few-shot transformer},
  author={Dong, Bowen and Zhou, Pan and Yan, Shuicheng and Zuo, Wangmeng},
  booktitle={European Conference on Computer Vision (ECCV)},
  pages={329--347},
  year={2022},
  organization={Springer}
}

@article{sun2023meta,
  title={Meta-AdaM: A meta-learned adaptive optimizer with momentum for few-shot learning},
  author={Sun, Siyuan and Gao, Hongyang},
  journal={Advances in Neural Information Processing Systems},
  volume={36},
  pages={65441--65455},
  year={2023}
}

@inproceedings{lin2023SMKD,
  title={Supervised masked knowledge distillation for few-shot transformers},
  author={Lin, Han and Han, Guangxing and Ma, Jiawei and Huang, Shiyuan and Lin, Xudong and Chang, Shih-Fu},
  booktitle={Proceedings of the IEEE/CVF Conference on Computer Vision and Pattern Recognition (CVPR)},
  pages={19649--19659},
  year={2023}
}

@article{jiang2024CPES,
  title={Class-relevant patch embedding selection for few-shot image classification},
  author={Jiang, Weihao and Cui, Haoyang and He, Kun},
  journal={Image and Vision Computing},
  pages={106031},
  year={2026},
  publisher={Elsevier}
}

@inproceedings{kang2021REnet,
  title={Relational embedding for few-shot classification},
  author={Kang, Dahyun and Kwon, Heeseung and Min, Juhong and Cho, Minsu},
  booktitle={Proceedings of the IEEE/CVF International Conference on Computer Vision (ICCV)},
  pages={8822--8833},
  year={2021}
}

@inproceedings{chen2021meta,
  title={{Meta-baseline: Exploring simple meta-learning for few-shot learning}},
  author={Chen, Yinbo and Liu, Zhuang and Xu, Huijuan and Darrell, Trevor and Wang, Xiaolong},
  booktitle={Proceedings of the IEEE/CVF International Conference on Computer Vision (ICCV)},
  pages={9062--9071},
  year={2021}
}

@techreport{welinder2010caltech,
  title={Caltech-UCSD Birds 200},
  author={Welinder, Peter and Branson, Steve and
  Mita, Takeshi and Wah, Catherine and
  Schroff, Florian and Belongie, Serge and
  Perona, Pietro},
  institution={California Institute of Technology},
  number={CNS-TR-2010-001},
  year={2010}
}

@inproceedings{krause2013StanfordCars,
  title={3d object representations for fine-grained categorization},
  author={Krause, Jonathan and Stark, Michael and Deng, Jia and Fei-Fei, Li},
  booktitle={Proceedings of the IEEE International Conference on Computer Vision Workshops (ICCVW)},
  pages={554--561},
  year={2013}
}

@article{hou2019cross,
  title={Cross attention network for few-shot classification},
  author={Hou, Ruibing and Chang, Hong and Ma, Bingpeng and Shan, Shiguang and Chen, Xilin},
  journal={Advances in Neural Information Processing Systems},
  volume={32},
  year={2019}
}

@inproceedings{tian2020rethinking,
  title={Rethinking few-shot image classification: a good embedding is all you need?},
  author={Tian, Yonglong and Wang, Yue and Krishnan, Dilip and Tenenbaum, Joshua B and Isola, Phillip},
  booktitle={European Conference on Computer Vision (ECCV)},
  pages={266--282},
  year={2020},
  organization={Springer}
}

@inproceedings{yang2024depth1,
  title={Depth anything: Unleashing the power of large-scale unlabeled data},
  author={Yang, Lihe and Kang, Bingyi and Huang, Zilong and Xu, Xiaogang and Feng, Jiashi and Zhao, Hengshuang},
  booktitle={Proceedings of the IEEE/CVF Conference on Computer Vision and Pattern Recognition (CVPR)},
  pages={10371--10381},
  year={2024}
}

@article{yang2024depth2,
  title={Depth anything v2},
  author={Yang, Lihe and Kang, Bingyi and Huang, Zilong and Zhao, Zhen and Xu, Xiaogang and Feng, Jiashi and Zhao, Hengshuang},
  journal={Advances in Neural Information Processing Systems},
  volume={37},
  pages={21875--21911},
  year={2024}
}

@inproceedings{caron2021DINO,
  title={Emerging properties in self-supervised vision transformers},
  author={Caron, Mathilde and Touvron, Hugo and Misra, Ishan and J{\'e}gou, Herv{\'e} and Mairal, Julien and Bojanowski, Piotr and Joulin, Armand},
  booktitle={Proceedings of the IEEE/CVF International Conference on Computer Vision (ICCV)},
  pages={9650--9660},
  year={2021}
}

@inproceedings{zhou2021ibot,
  title={iBOT: Image BERT Pre-Training with Online Tokenizer},
  author={Jinghao Zhou and Chen Wei and Huiyu Wang and Wei Shen and Cihang Xie and Alan Yuille and Tao Kong},
  booktitle={International Conference on Learning Representations (ICLR)},
  year={2022}
}

@inproceedings{fu2020depthfsl,
  title={Depth guided adaptive meta-fusion network for few-shot video recognition},
  author={Fu, Yuqian and Zhang, Li and Wang, Junke and Fu, Yanwei and Jiang, Yu-Gang},
  booktitle={Proceedings of the ACM International Conference on Multimedia},
  pages={1142--1151},
  year={2020}
}

@inproceedings{cumplido2025slot,
  title={Slot attention-based feature filtering for few-shot learning},
  author={Cumplido, Javier R{\'o}denas and Aguilar, Eduardo and Radeva, Petia},
  booktitle={Proceedings of the IEEE/CVF Conference on Computer Vision and Pattern Recognition Workshops (CVPRW)},
  pages={30--40},
  year={2025}
}

@inproceedings{cumplido2025stochastic,
  title={Stochastic-based Patch Filtering for Few-Shot Learning},
  author={Cumplido, Javier R{\'o}denas and Aguilar, Eduardo and Radeva, Petia},
  booktitle={Proceedings of the IEEE/CVF Conference on Computer Vision and Pattern Recognition Workshops (CVPRW)},
  pages={440--449},
  year={2025}
}

@inproceedings{zhang2024simple,
  title={Simple semantic-aided few-shot learning},
  author={Zhang, Hai and Xu, Junzhe and Jiang, Shanlin and He, Zhenan},
  booktitle={Proceedings of the IEEE/CVF Conference on Computer Vision and Pattern Recognition (CVPR)},
  pages={28588--28597},
  year={2024}
}

@inproceedings{li2024fewvs,
  title={{FewVS: A vision-semantics integration framework for few-shot image classification}},
  author={Li, Zhuoling and Wang, Yong and Li, Kaitong},
  booktitle={Proceedings of the ACM International Conference on Multimedia},
  pages={1341--1350},
  year={2024}
}

@inproceedings{liu2025envisioning,
  title={Envisioning class entity reasoning by large language models for few-shot learning},
  author={Liu, Mushui and Wu, Fangtai and Li, Bozheng and Lu, Ziqian and Yu, Yunlong and Li, Xi},
  booktitle={Proceedings of the AAAI Conference on Artificial Intelligence (AAAI)},
  volume={39},
  number={18},
  pages={18906--18914},
  year={2025}
}

@inproceedings{ren2018tieredImageNet,
  title={Meta-learning for semi-supervised few-shot classification},
  author={Ren, Mengye and Triantafillou, Eleni and Ravi, Sachin and Snell, Jake and Swersky, Kevin and Tenenbaum, Joshua B and Larochelle, Hugo and Zemel, Richard S},
  booktitle={International Conference on Learning Representations (ICLR)},
  year={2018}
}

@inproceedings{
bertinetto2018CifarFS,
title={Meta-learning with differentiable closed-form solvers},
author={Luca Bertinetto and Joao F. Henriques and Philip Torr and Andrea Vedaldi},
booktitle={International Conference on Learning Representations (ICLR)},
year={2019}
}

@article{oreshkin2018FC100,
  title={{Tadam: Task-dependent adaptive metric for improved few-shot learning}},
  author={Oreshkin, Boris and Rodr{\'\i}guez L{\'o}pez, Pau and Lacoste, Alexandre},
  journal={Advances in Neural Information Processing Systems},
  volume={31},
  year={2018}
}

@inproceedings{lee2019MetaOptNet,
  title={Meta-learning with differentiable convex optimization},
  author={Lee, Kwonjoon and Maji, Subhransu and Ravichandran, Avinash and Soatto, Stefano},
  booktitle={Proceedings of the IEEE/CVF Conference on Computer Vision and Pattern Recognition (CVPR)},
  pages={10657--10665},
  year={2019}
}

@inproceedings{dosovitskiy2020ViT,
  title={An image is worth 16x16 words: Transformers for image recognition at scale},
  author={Dosovitskiy, Alexey and Beyer, Lucas and Kolesnikov, Alexander and Weissenborn, Dirk and Zhai, Xiaohua and Unterthiner, Thomas and Dehghani, Mostafa and Minderer, Matthias and Heigold, Georg and Gelly, Sylvain and others},
  booktitle={International Conference on Learning Representations (ICLR)},
  year={2021}
}

@inproceedings{diederik2014adam,
  title={ADAM: A method for stochastic optimization.},
  author={Kingma, Diederik P. and Ba, Jimmy},
  booktitle={International Conference on Learning Representations (ICLR)},
  year={2015}
}

@article{elfwing2018silu,
  title={Sigmoid-weighted linear units for neural network function approximation in reinforcement learning},
  author={Elfwing, Stefan and Uchibe, Eiji and Doya, Kenji},
  journal={Neural Networks},
  volume={107},
  pages={3--11},
  year={2018},
  publisher={Elsevier}
}

@article{ulyanov2016instancenorm,
  title={Instance normalization: The missing ingredient for fast stylization},
  author={Ulyanov, Dmitry and Vedaldi, Andrea and Lempitsky, Victor},
  journal={arXiv preprint arXiv:1607.08022},
  year={2016}
}

@inproceedings{Oh2021BOILTR,
  title={BOIL: Towards Representation Change for Few-shot Learning},
  author={Jaehoon Oh and Hyungjun Yoo and ChangHwan Kim and Seyoung Yun},
  booktitle={International Conference on Learning Representations (ICLR)},
  year={2021}
}

@article{zhou2024metaFreq,
  title={Meta-exploiting frequency prior for cross-domain few-shot learning},
  author={Zhou, Fei and Wang, Peng and Zhang, Lei and Chen, Zhenghua and Wei, Wei and Ding, Chen and Lin, Guosheng and Zhang, Yanning},
  journal={Advances in Neural Information Processing Systems},
  volume={37},
  pages={116783--116814},
  year={2024}
}

@inproceedings{chen2021sPlusf,
  title={Few-shot learning by integrating spatial and frequency representation},
  author={Chen, Xiangyu and Wang, Guanghui},
  booktitle={Proceedings of Conference on Robots and Vision (CRV)},
  pages={49--56},
  year={2021},
  organization={IEEE}
}

@article{zheng2023diffkendall,
  title={DiffKendall: A Novel Approach for Few-Shot Learning with Differentiable Kendall's Rank Correlation},
  author={Zheng, Kaipeng and Zhang, Huishuai and Huang, Weiran},
  journal={Advances in Neural Information Processing Systems},
  volume={36},
  pages={49403--49415},
  year={2023}
}

@inproceedings{xie2022DeepBDC,
  title={Joint distribution matters: Deep brownian distance covariance for few-shot classification},
  author={Xie, Jiangtao and Long, Fei and Lv, Jiaming and Wang, Qilong and Li, Peihua},
  booktitle={Proceedings of the IEEE/CVF Conference on Computer Vision and Pattern Recognition (CVPR)},
  pages={7972--7981},
  year={2022}
}

@InProceedings{Marigold2024,
    author    = {Ke, Bingxin and Obukhov, Anton and Huang, Shengyu and Metzger, Nando and Daudt, Rodrigo Caye and Schindler, Konrad},
    title     = {Repurposing Diffusion-Based Image Generators for Monocular Depth Estimation},
    booktitle = {Proceedings of the IEEE/CVF Conference on Computer Vision and Pattern Recognition (CVPR)},
    month     = {June},
    year      = {2024},
    pages     = {9492-9502}
}

@inproceedings{beit,
title={{BEiT}: {BERT} Pre-Training of Image Transformers},
author={Hangbo Bao and Li Dong and Songhao Piao and Furu Wei},
booktitle={International Conference on Learning Representations (ICLR)},
year={2022}
}

@article{
oquab2024dinov2,
title={{DINO}v2: Learning Robust Visual Features without Supervision},
author={Maxime Oquab and Timoth{\'e}e Darcet and Th{\'e}o Moutakanni and Huy V. Vo and Marc Szafraniec and Vasil Khalidov and Pierre Fernandez and Daniel HAZIZA and Francisco Massa and Alaaeldin El-Nouby and Mido Assran and Nicolas Ballas and Wojciech Galuba and Russell Howes and Po-Yao Huang and Shang-Wen Li and Ishan Misra and Michael Rabbat and Vasu Sharma and Gabriel Synnaeve and Hu Xu and Herve Jegou and Julien Mairal and Patrick Labatut and Armand Joulin and Piotr Bojanowski},
journal={Transactions on Machine Learning Research},
issn={2835-8856},
year={2024}
}

@InProceedings{HyRSM_cvpr2022,
    author    = {Wang, Xiang and Zhang, Shiwei and Qing, Zhiwu and Tang, Mingqian and Zuo, Zhengrong and Gao, Changxin and Jin, Rong and Sang, Nong},
    title     = {Hybrid Relation Guided Set Matching for Few-Shot Action Recognition},
    booktitle = {Proceedings of the IEEE/CVF Conference on Computer Vision and Pattern Recognition (CVPR)},
    month     = {June},
    year      = {2022},
    pages     = {19948-19957}
}

@InProceedings{Molo_cvpr2023,
    author    = {Wang, Xiang and Zhang, Shiwei and Qing, Zhiwu and Gao, Changxin and Zhang, Yingya and Zhao, Deli and Sang, Nong},
    title     = {MoLo: Motion-Augmented Long-Short Contrastive Learning for Few-Shot Action Recognition},
    booktitle = {Proceedings of the IEEE/CVF Conference on Computer Vision and Pattern Recognition (CVPR)},
    month     = {June},
    year      = {2023},
    pages     = {18011-18021}
}

@InProceedings{Hausler_2021_CVPR,
    author    = {Hausler, Stephen and Garg, Sourav and Xu, Ming and Milford, Michael and Fischer, Tobias},
    title     = {Patch-NetVLAD: Multi-Scale Fusion of Locally-Global Descriptors for Place Recognition},
    booktitle = {Proceedings of the IEEE/CVF Conference on Computer Vision and Pattern Recognition (CVPR)},
    month     = {June},
    year      = {2021},
    pages     = {14141-14152}
}

@InProceedings{Dekel_2015_CVPR,
author = {Dekel, Tali and Oron, Shaul and Rubinstein, Michael and Avidan, Shai and Freeman, William T.},
title = {Best-Buddies Similarity for Robust Template Matching},
booktitle = {Proceedings of the IEEE Conference on Computer Vision and Pattern Recognition (CVPR)},
month = {June},
year = {2015}
}

@inproceedings{he2022MaskedEncoder,
  title={Masked autoencoders are scalable vision learners},
  author={He, Kaiming and Chen, Xinlei and Xie, Saining and Li, Yanghao and Doll{\'a}r, Piotr and Girshick, Ross},
  booktitle={Proceedings of the IEEE/CVF Conference on Computer Vision and Pattern Recognition (CVPR)},
  pages={16000--16009},
  year={2022}
}

@inproceedings{rizve2021IE,
  title={Exploring complementary strengths of invariant and equivariant representations for few-shot learning},
  author={Rizve, Mamshad Nayeem and Khan, Salman and Khan, Fahad Shahbaz and Shah, Mubarak},
  booktitle={Proceedings of the IEEE/CVF Conference on Computer Vision and Pattern Recognition (CVPR)},
  pages={10836--10846},
  year={2021}
}

@inproceedings{wu2021TPMN,
  title={Task-aware part mining network for few-shot learning},
  author={Wu, Jiamin and Zhang, Tianzhu and Zhang, Yongdong and Wu, Feng},
  booktitle={Proceedings of the IEEE/CVF International Conference on Computer Vision (ICCV)},
  pages={8433--8442},
  year={2021}
}

@inproceedings{chen2021PSST,
  title={Pareto self-supervised training for few-shot learning},
  author={Chen, Zhengyu and Ge, Jixie and Zhan, Heshen and Huang, Siteng and Wang, Donglin},
  booktitle={Proceedings of the IEEE/CVF Conference on Computer Vision and Pattern Recognition (CVPR)},
  pages={13663--13672},
  year={2021}
}

@inproceedings{qi2021OM,
  title={Transductive few-shot classification on the oblique manifold},
  author={Qi, Guodong and Yu, Huimin and Lu, Zhaohui and Li, Shuzhao},
  booktitle={Proceedings of the IEEE/CVF International Conference on Computer Vision (ICCV)},
  pages={8412--8422},
  year={2021}
}

@inproceedings{zhang2021Meta_QDA,
  title={Shallow bayesian meta learning for real-world few-shot recognition},
  author={Zhang, Xueting and Meng, Debin and Gouk, Henry and Hospedales, Timothy M},
  booktitle={Proceedings of the IEEE/CVF International Conference on Computer Vision (ICCV)},
  pages={651--660},
  year={2021}
}

@inproceedings{gidaris2019CCrot,
  title={Boosting few-shot visual learning with self-supervision},
  author={Gidaris, Spyros and Bursuc, Andrei and Komodakis, Nikos and P{\'e}rez, Patrick and Cord, Matthieu},
  booktitle={Proceedings of the IEEE/CVF International Conference on Computer Vision (ICCV)},
  pages={8059--8068},
  year={2019}
}

@inproceedings{liu2022dmn4,
  title={Dmn4: Few-shot learning via discriminative mutual nearest neighbor neural network},
  author={Liu, Yang and Zheng, Tu and Song, Jie and Cai, Deng and He, Xiaofei},
  booktitle={Proceedings of the AAAI conference on artificial intelligence (AAAI)},
  volume={36},
  number={2},
  pages={1828--1836},
  year={2022}
}

@inproceedings{kirillov2023sam,
  title={Segment anything},
  author={Kirillov, Alexander and Mintun, Eric and Ravi, Nikhila and Mao, Hanzi and Rolland, Chloe and Gustafson, Laura and Xiao, Tete and Whitehead, Spencer and Berg, Alexander C and Lo, Wan-Yen and others},
  booktitle={Proceedings of the IEEE/CVF international conference on computer vision (ICCV)},
  pages={4015--4026},
  year={2023}
}

@inproceedings{carion2026sam3,
  title     = {SAM 3: Segment Anything with Concepts},
  author    = {Carion, Nicolas and Gustafson, Laura and Hu, Yuan-Ting and Debnath, Shoubhik and Hu, Ronghang and Suris Coll-Vinent, Didac and Ryali, Chaitanya and Alwala, Kalyan Vasudev and Khedr, Haitham and Huang, Andrew and Lei, Jie and Ma, Tengyu and Guo, Baishan and Kalla, Arpit and Marks, Markus and Greer, Joseph and Wang, Meng and Sun, Peize and R{\"a}dle, Roman and Afouras, Triantafyllos and others},
  booktitle = {International Conference on Learning Representations (ICLR)},
  year      = {2026}
}

@article{ren2024groundedsam,
  title={Grounded sam: Assembling open-world models for diverse visual tasks},
  author={Ren, Tianhe and Liu, Shilong and Zeng, Ailing and Lin, Jing and Li, Kunchang and Cao, He and Chen, Jiayu and Huang, Xinyu and Chen, Yukang and Yan, Feng and others},
  journal={arXiv preprint arXiv:2401.14159},
  year={2024}
}

@inproceedings{radford2021learning,
  title={Learning transferable visual models from natural language supervision},
  author={Radford, Alec and Kim, Jong Wook and Hallacy, Chris and Ramesh, Aditya and Goh, Gabriel and Agarwal, Sandhini and Sastry, Girish and Askell, Amanda and Mishkin, Pamela and Clark, Jack and others},
  booktitle={Proceedings of International Conference on Machine Learning (ICML)},
  pages={8748--8763},
  year={2021}
}
}

\maketitlesupplementary
\appendix

This document provides supplementary material for the main manuscript, organized as follows. Section~\ref{sec:dataset} offers a brief recap of the evaluation datasets. Section~\ref{sec:pretraining} details the implementation and analysis of our RGB-D pre-training strategy. Section~\ref{sec:FSL} presents additional quantitative and qualitative results in the few-shot learning stage. Section~\ref{sec:ablation} includes more ablation studies absent from the main paper due to space constraints. Section~\ref{sec:discussion} provides the discussion of DAFIC and future work. Finally, Section~\ref{sec:depth_potential} investigates the integration of depth cues into several existing few-shot image classification (FSIC) methods, demonstrating the broader utility of depth priors for FSIC domain.

\section{Dataset}
\label{sec:dataset}
All experiments are conducted on four standard FSIC benchmarks following most of the prior FSIC works: MiniImageNet \cite{vinyals2016MatchingNet}, TieredImageNet \cite{ren2018tieredImageNet}, CIFAR-FS \cite{bertinetto2018CifarFS}, and FC100 \cite{oreshkin2018FC100}. 
\vspace{-2mm}

\begin{table}[!ht]
\scriptsize
\centering
\caption{Commonly adopted FSIC benchmark datasets in prior works.}
\begin{tabular}{@{}lcc@{}}
\toprule
\textbf{Dataset} & \textbf{\# Images} & \textbf{Train/Val/Test} \\ 
\midrule
MiniImageNet~\cite{vinyals2016MatchingNet} & $\sim$60{,}000   & 64/16/20 \\
TieredImageNet~\cite{ren2018tieredImageNet} & $\sim$779{,}165 & 351/97/160 \\
CIFAR-FS~\cite{bertinetto2018CifarFS} & $\sim$60{,}000     & 64/16/20 \\
FC100~\cite{oreshkin2018FC100} & $\sim$60{,}000         & 60/20/20 \\
\bottomrule
\end{tabular}
\vspace{-5mm}
\label{Tab:dataset}
\end{table}

\begin{table}[t]
\centering
\caption{Summary of the RGB-D Pre-training Configuration}
\renewcommand{\arraystretch}{1.20}
\newcommand{\dashrule}{\addlinespace[0.5ex]\hdashline[1.5pt/2pt]\addlinespace[0.5ex]}
\resizebox{0.85\linewidth}{!}{
\begin{tabular}{p{0.30\linewidth} p{0.60\linewidth}}
\toprule

\multicolumn{1}{c}{\textbf{Category}} &
\multicolumn{1}{c}{\textbf{Configuration}} \\

\midrule

\centering\textbf{Backbone} &
ViT-Small with $16{\times}16$ patches; 8192-d projection head shared by class and patch tokens (student/teacher). \\
\dashrule
\centering\textbf{Optimization} &
AdamW optimizer; cosine LR from $5\times10^{-4}$ to $1\times10^{-6}$;  
weight decay 0.04~$\rightarrow$~0.4; 10-epoch warm-up. \\
\dashrule
\centering\textbf{Teacher EMA} &
EMA momentum: 0.996; teacher temperature increases from 0.04 to 0.07 during the first 30 epochs. \\
\dashrule
\centering\textbf{Augmentation} &
RGB-D multi-crop strategy:  
2 global crops ($224{\times}224$, scale [0.25,1.0]) and  
10 local crops ($96{\times}96$, scale [0.05,0.25]);  
RGB-depth synchronized crop/flip;  
per-channel RGB and depth normalization. \\
\dashrule
\centering\textbf{MIM} &
Block-wise masking on global crops (student network);  
masking ratio sampled from [0.0, 0.3] with variance [0.0, 0.2]. \\
\dashrule
\centering\textbf{Computation} &
Distributed training with DDP; 3$\times$A100 (40GB) GPUs; batch size 384; pre-training for 1,600 epochs (800 for TieredImageNet); checkpoint selected at epoch 1,000 (500 for TieredImageNet). \\

\bottomrule
\end{tabular}}
\label{Tab:pretraining}
\end{table}

\begin{figure*}[!t]
    \centering
    \begin{minipage}{0.49\linewidth}
        \centering
        \includegraphics[width=\linewidth]{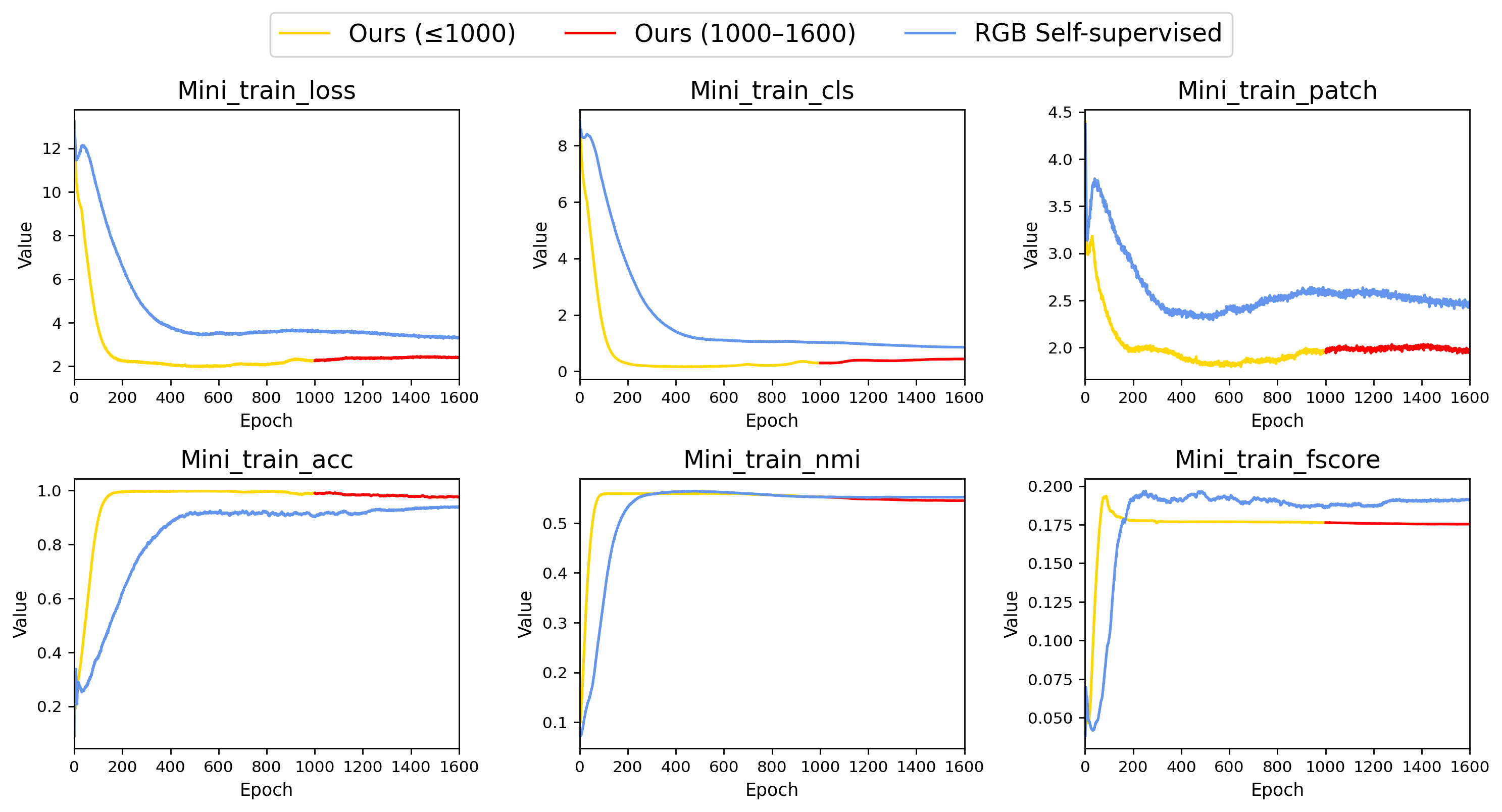}
        \caption*{(a) MiniImageNet}
    \end{minipage}
    \hfill
    \begin{minipage}{0.49\linewidth}
        \centering
        \includegraphics[width=\linewidth]{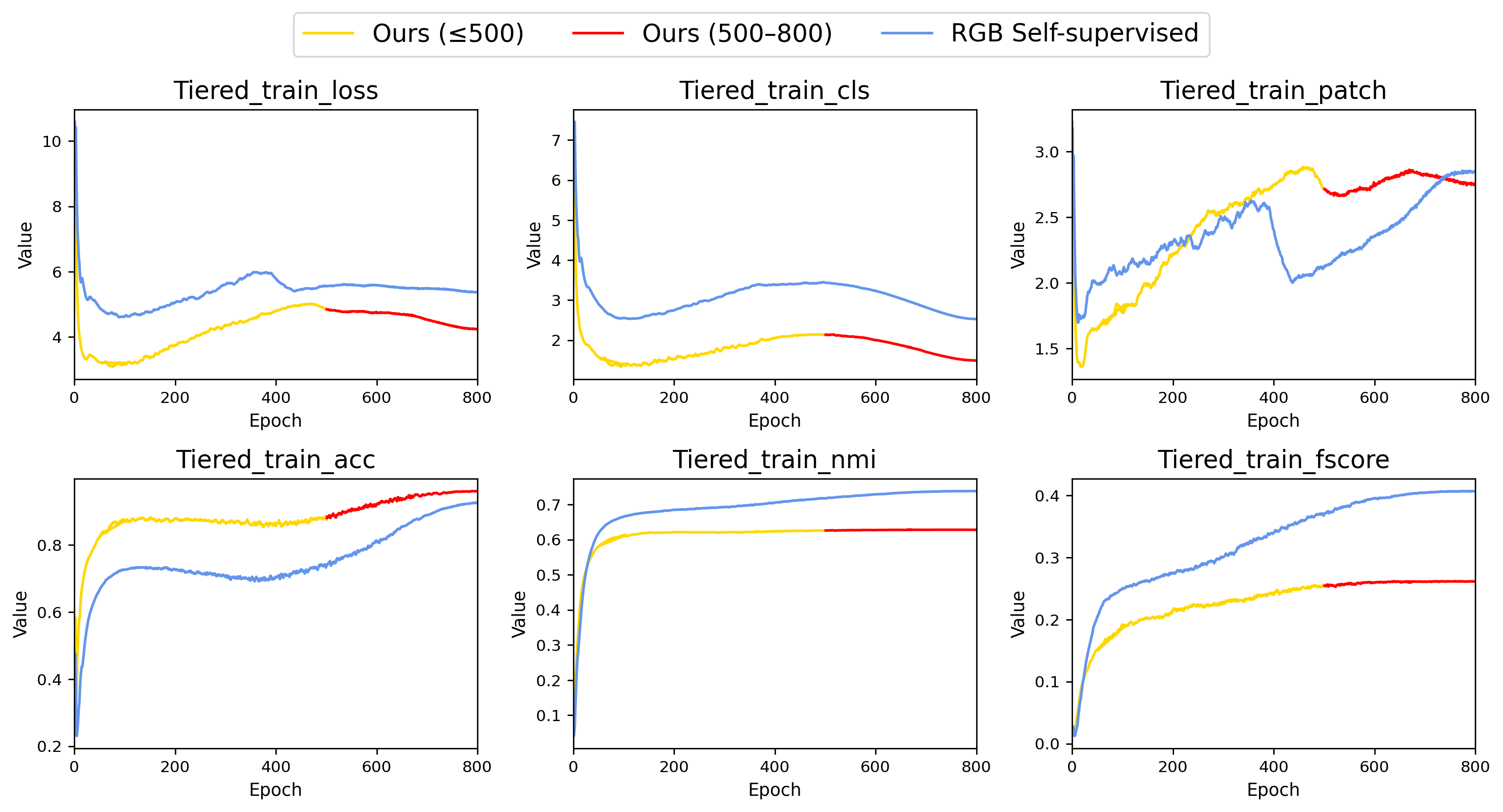}
        \caption*{(b) TieredImageNet}
    \end{minipage}

    \vspace{0.3cm}

    \begin{minipage}{0.49\linewidth}
        \centering
        \includegraphics[width=\linewidth]{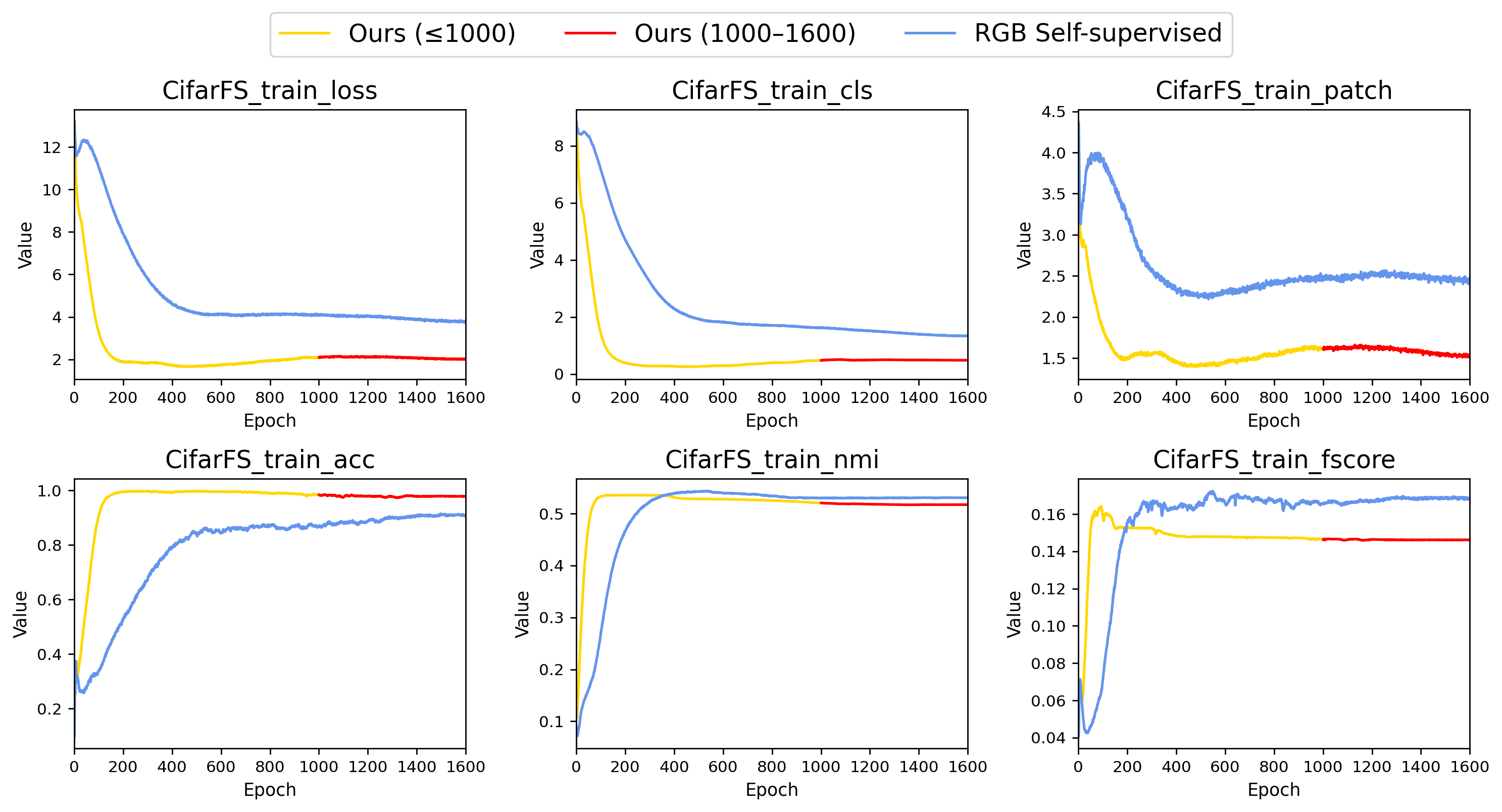}
        \caption*{(c) CIFAR-FS}
    \end{minipage}
    \hfill
    \begin{minipage}{0.49\linewidth}
        \centering
        \includegraphics[width=\linewidth]{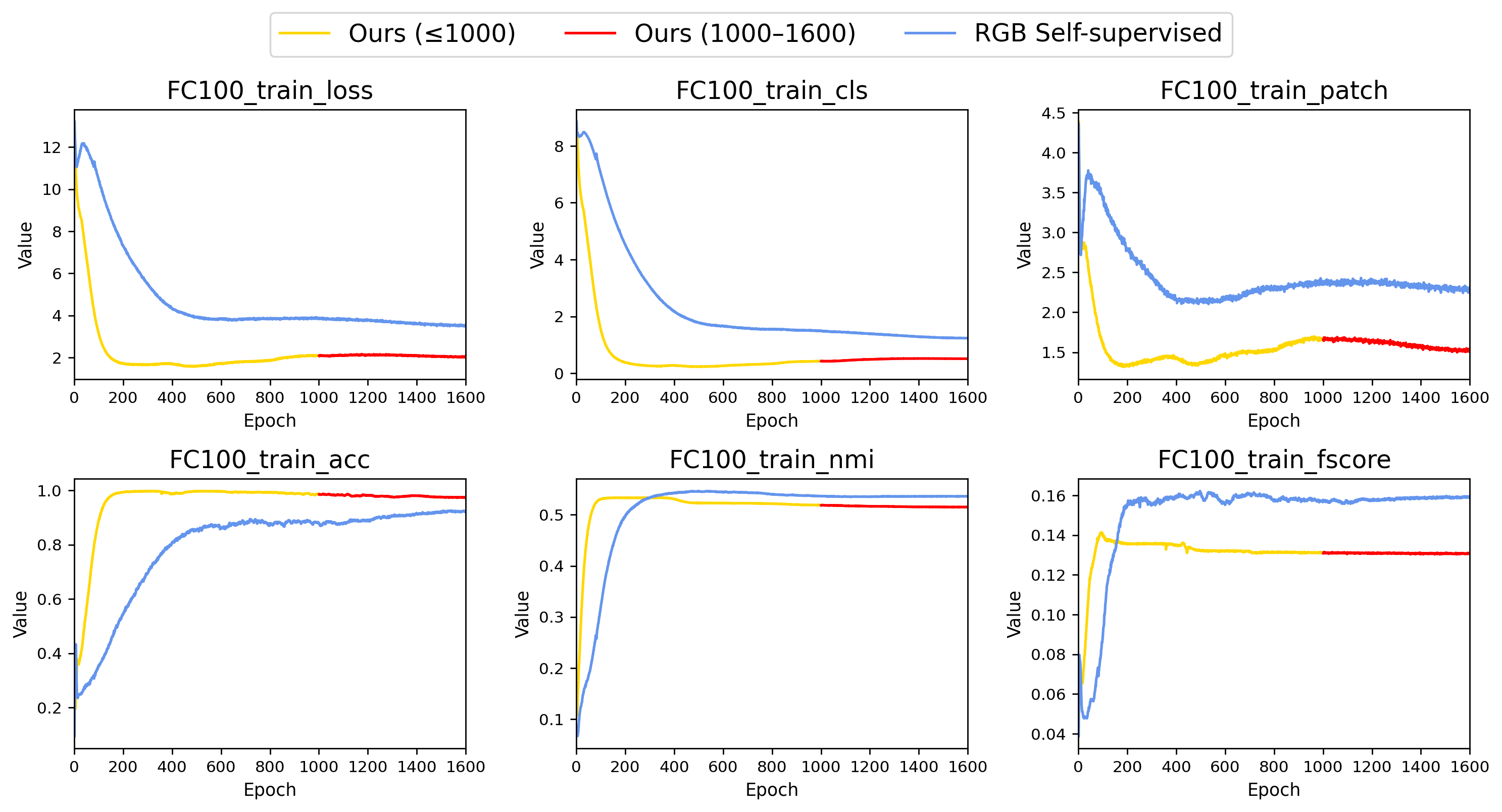}
        \caption*{(d) FC100}
    \end{minipage}

    \caption{
     Comparison between our RGB-D and RGB-only self-supervised pre-training across four datasets, using six metrics to monitor the training dynamics. RGB-D pre-training generally achieves lower total and class-level losses, converges faster, and yields higher training accuracy than RGB-only pre-training, while its clustering metrics (NMI and F-score) are generally lower. The yellow and red curves denote the training phases before and after the selected checkpoint used for downstream few-shot classification.
    }
    \label{fig:pretrain_metrics}
\end{figure*}

\section{Implementation Details for Pre-training}
\label{sec:pretraining}
\subsection{RGB-D Pre-training Setups}
We pretrain the Vision Transformer (ViT) backbone and projection heads inspired by the prior works of FewTURE~\cite{hiller2022Fewtrue} and CPEA \cite{hao2023CPEA}. Table \ref{Tab:pretraining} provides the configuration summary of our RGB-D pretraining. Specifically, we adopt a ViT-Small architecture with a $16{\times}16$ patch size, using a shared 8192-dimensional projection head for both the class token and the patch tokens in the student and teacher networks, ensuring the global and local representations in a common embedding space. 
The model is optimized using the AdamW optimizer under a cosine learning-rate schedule, starting from $5{\times}10^{-4}$ and decaying to $1{\times}10^{-6}$, while the weight decay increases from 0.04 to 0.4. 
A 10-epoch linear warm-up is applied at the beginning of training. The teacher network is updated with an exponential moving-average (EMA) momentum of 0.996, and its temperature is linearly increased from 0.04 to 0.07 during the first 30 epochs to stabilize optimization. 
For data augmentation, we employ the multi-crop strategy~\cite{caron2021DINO} in an RGB-D setting, generating two global crops (size $224{\times}224$, scale range [0.25, 1.0]) and ten local crops (size $96{\times}96$, scale range [0.05, 0.25]). We also apply jointly aligned RGB–depth cropping/flipping and separate normalization for each channel.
Masked image modeling (MIM) \cite{beit, he2022MaskedEncoder, zhou2021ibot} is applied to global crops in the student branch using block-wise masking, where the masking ratio is uniformly sampled from [0.0, 0.3] with a variance of [0.0, 0.2]. We pre-train the RGB-D model for 1600 epochs on MiniImageNet, CIFAR-FS, and FC100, and for 800 epochs on TieredImageNet. For downstream few-shot classification, we use the checkpoint at epoch 1000 for MiniImageNet, CIFAR-FS, and FC100, and at epoch 500 for TieredImageNet. For comparison, the baseline pre-training uses four A100 GPUs with a batch size of 512 and 1,600/800 epochs.

\subsection{Pre-training Comparison Analysis}
In Figure~\ref{fig:pretrain_metrics}, we compare our RGB-D self-supervised pre-training with the RGB-only self-supervised pre-training baselines~\cite{hiller2022Fewtrue,hao2023CPEA} across these four benchmarks using six monitoring metrics. Empirically, we observe that RGB-D pre-training converges faster and reaches lower total and class-level losses, with earlier stabilization in several metrics. 
Therefore, we (1) follow the baseline setup to pretrain for 1600 epochs on most datasets and 800 epochs on TieredImageNet, 
but with a smaller total batch size (384 vs.\ 512) due to our computational constraints, 
and (2) adopt the checkpoints from 1000 epochs of most datasets and 500 epochs of TieredImageNet as the RGB-D pretrained backbones for subsequent few-shot classification.

We follow \cite{zhou2021ibot,hiller2022Fewtrue} to track the training dynamics throughout pre-training (up to 1600 epochs for MiniImageNet, CIFAR-FS, and FC100, and up to 800 epochs for TieredImageNet). The tracked metrics include:\\ 
(1) \textbf{Optimization objectives.} The total loss (\textit{train\_loss}), class-level distillation loss (\textit{train\_cls}), and patch-level MIM loss (\textit{train\_patch}) together characterize the overall training objective and the convergence behavior of both global (CLS-token) and local (patch-token) representations; (2) \textbf{Cross-view consistency.} The training accuracy (\textit{train\_acc}), reflecting the alignment stability between teacher and student predictions across different augmented views; (3) \textbf{Global clustering quality.} The normalized mutual information (\textit{train\_nmi}) and Fowlkes–Mallows score (\textit{train\_fscore}) are used to
assess how well the learned representations are semantically clustered by comparing the teacher’s codebook assignments with the ground-truth class labels.

Overall, RGB-D pre-training consistently yields faster convergence, lower \textit{train\_loss} and \textit{train\_cls}, and higher \textit{train\_acc} across the four benchmarks. These trends suggest that the additional depth signal facilitates teacher--student alignment during self-supervised MIM pre-training. For \textit{train\_patch}, RGB-D pretraining is generally competitive and often lower in the early stage, although the advantage is less consistent on TieredImageNet. In contrast, the clustering metrics (\textit{train\_nmi} and \textit{train\_fscore}) are generally lower than those of RGB-only pre-training. This indicates that RGB-D pre-training does not necessarily improve semantic clustering under these monitoring metrics, but instead appears to favor representations that are more effective for the downstream few-shot classification task, as evidenced by the final FSIC performance reported in the main paper, Table~\ref{tab:table_fulllist1} and Table~\ref{tab:table_fulllist2} in this document.

\section{Few-shot Learning with DAFIC}
\label{sec:FSL}

After RGB-D pre-training, the original projection head with an output dimension of 8,192 is removed and replaced with a new projection head of dimension 384, matching the embedding size of the ViT-Small backbone used in our framework. The new projection head is implemented as a two-layer multi-layer perceptron (MLP) with linear layers and GELU activations. We then train the full DAFIC framework, including the Depth-guided Geometric Adaptation (DGA) module and the Spatial Bidirectional Patch Matching (SBiM) classifier, under a meta-training paradigm to promote cross-task generalization. This training protocol follows most prior metric-based FSIC methods \cite{ye2020FEAT, hiller2022Fewtrue, hao2023CPEA, cumplido2025slot}. In this stage, both training and inference are conducted on a single NVIDIA V100 GPU.

For completeness, we briefly clarify \textbf{why the bidirectional matching in SBiM is not redundant}, although both directions are derived from the same similarity matrix. The query-to-support score performs row-wise max-pooling, measuring whether each query patch can find strong evidence in the support class, whereas the support-to-query score performs column-wise max-pooling, whether the support patches are sufficiently covered by the query image. Therefore, the two directions capture complementary aspects of dense correspondence. Using only the forward direction may lead to many-to-one matches, where multiple query patches repeatedly align with a few salient support patches and thus overestimate similarity. The backward direction helps alleviate this issue by encouraging more structurally consistent query-support alignment.

\textbf{Quantitative Comparison with Prior FSIC Methods.} DAFIC consistently outperforms the baseline and achieves strong performance compared with prior FSIC methods on both 5-way 1-shot and 5-way 5-shot tasks across all datasets, as summarized in Table~\ref{tab:table_fulllist1} and Table~\ref{tab:table_fulllist2}.

\textbf{Qualitative Visualizations Compared to the Baseline.} Beyond the quantitative gains, DAFIC focuses on class-relevant regions more reliably than the baseline across diverse FSIC scenarios, including distracting backgrounds, multiple entities, and cluttered scenes. These visualizations suggest that our DAFIC alleviates the ambiguity introduced by relying solely on RGB appearance and encourages more class-focused attention, as shown in Figure~\ref{fig:patchclass_alignment}.

\vspace{4mm}
\begin{figure}[h]
    \centering
    \includegraphics[width=0.45\textwidth]{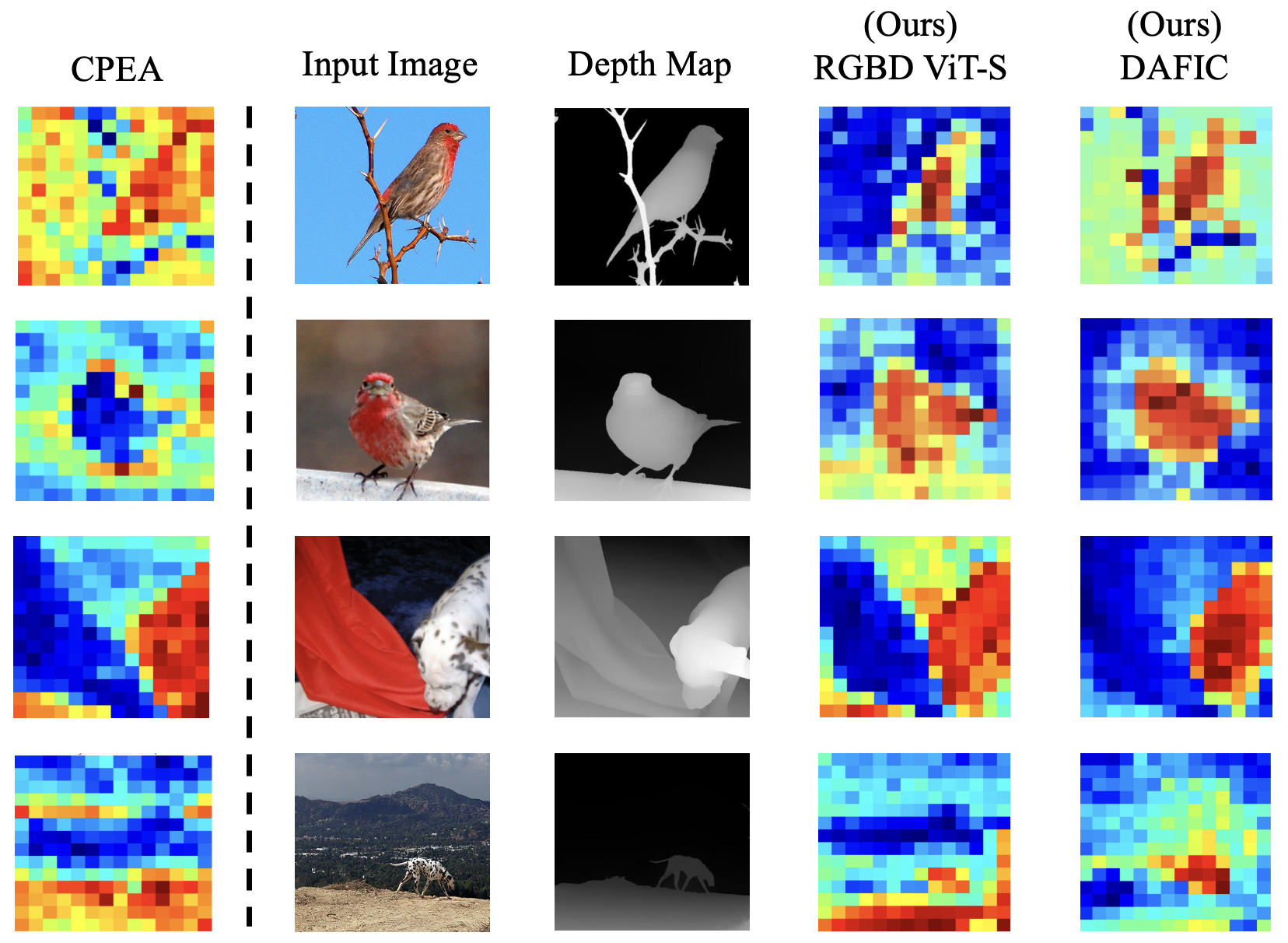}
    \includegraphics[width=0.45\textwidth]{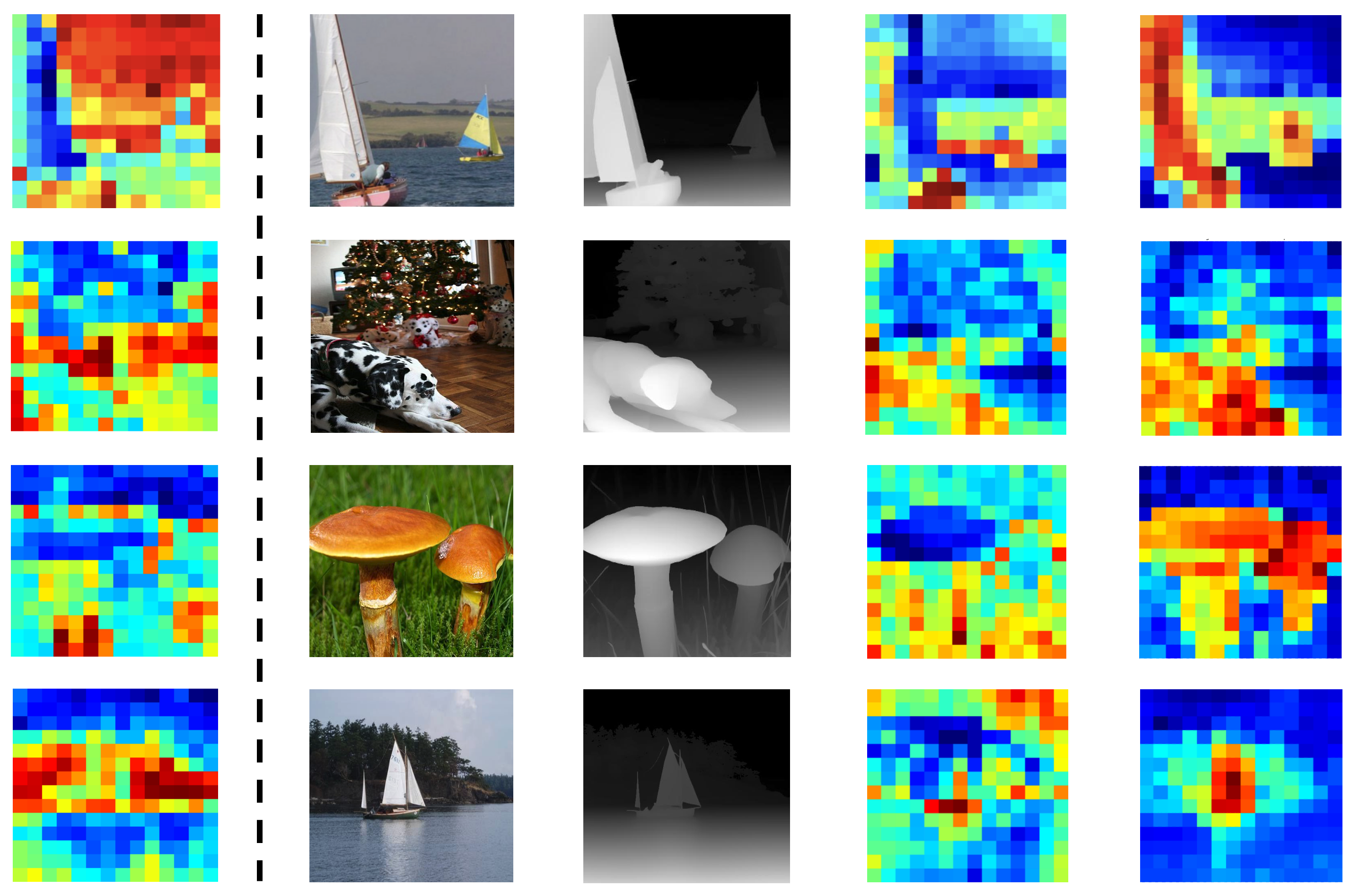}
    \includegraphics[width=0.45\textwidth]{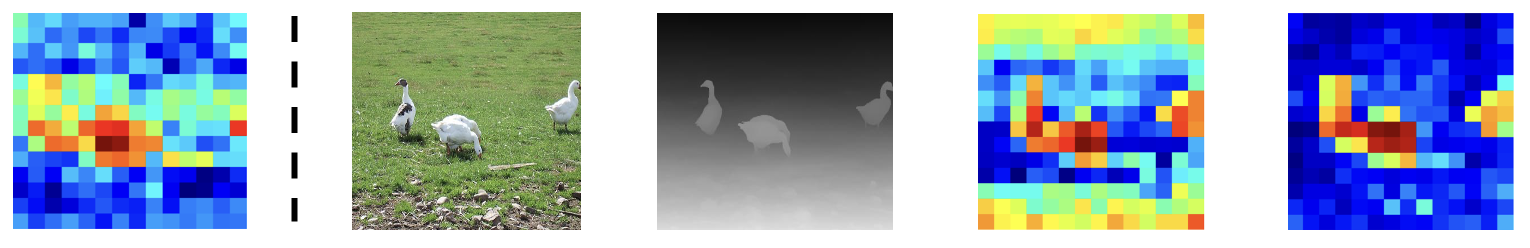}
    \caption{More visualizations of patch-to-class semantic alignment between baseline CPEA~\cite{hao2023CPEA} and DAFIC. Warmer colors (towards red) indicate stronger class relevance, whereas cooler colors (towards blue) denote weaker or non-relevant regions. As shown, the RGB-D ViT backbone alone already yields more class-consistent responses than the baseline. After applying the full DAFIC framework, the patch-to-class alignment is further improved and becomes more precise.}
    \label{fig:patchclass_alignment}
\end{figure}

\begin{table*}[t]
    \centering
    \scriptsize
    \caption{More comprehensive results (\%) on MiniImageNet and TieredImageNet. The ± shows 95\% confidence intervals. The best and second-best results are shown in \textbf{bold} and \underline{underlined}.}
     \resizebox{0.99\linewidth}{!}{
     \begin{tabular}{c c l c c c c c c}
    \toprule
         &\multirow{2}{*}{\textbf{Method}} &
         \multirow{2}{*}{\textbf{Venue}} & 
         \multirow{2}{*}{\textbf{Backbone}} &
         \multirow{2}{*}{\textbf{$\approx$ \# Params}} &
         \multicolumn{2}{c}{\textbf{MiniImageNet}} &
         \multicolumn{2}{c}{\textbf{TieredImageNet}}\\   \cmidrule(lr){6-7}\cmidrule(lr){8-9}&
         \multirow{2}{*}{} &
         \multirow{2}{*}{} & 
         \multirow{2}{*}{} & 
         \multirow{2}{*}{} & 
         \textbf{5-way 1-shot} & 
         \textbf{5-way 5-shot}  & 
         \textbf{5-way 1-shot} & 
         \textbf{5-way 5-shot}\\ 
        \midrule
        &MatchNet \cite{vinyals2016MatchingNet} & NeurIPS'16 & ResNet-12 & 12.4M & 65.64 ± 0.20 & 78.72 ± 0.15 & 68.50 ± 0.92 & 80.60 ± 0.71 \\
        &ProtoNet \cite{snell2017ProtoNet} & NeurIPS'17  & ResNet-12 & 12.4M & 62.29 ± 0.33 & 79.46 ± 0.48 & 68.25 ± 0.23 & 84.01 ± 0.56 \\
        &MAML \cite{finn2017model} & ICML'17  & ResNet-12 & 12.4M & 49.24 ± 0.21 & 58.05 ± 0.10 & 67.92 ± 0.17 & 72.41 ± 0.20 \\
        &TADAM \cite{oreshkin2018FC100} & NeurIPS'18 & ResNet-12 & 12.4M &  58.50 ± 0.30 & 76.70 ± 0.30 & 62.13 ± 0.31 & 81.92 ± 0.30 \\
        &LEO \cite{rusu2018meta} & ICLR'19 & WRN-28-10 & 36.5M & 61.76 ± 0.08 & 77.59 ± 0.12 & 66.33 ± 0.05 & 81.44 ± 0.09 \\
        &DN4 \cite{Li2019DN4} & CVPR'19 & ResNet-12 & 12.4M & 56.93 ± 0.62 & 73.97 ± 0.49 & 66.25 ± 0.75 & 82.62 ± 0.51 \\
        &CAN \cite{hou2019cross} & NeurIPS'19 & ResNet-12 & 12.4M & 63.85 ± 0.48 & 79.44 ± 0.34 & 69.89 ± 0.51 & 84.23 ± 0.37 \\
        &CTM \cite{li2019CTM} & CVPR'19 & ResNet-12 & 12.4M & 64.12 ± 0.82 & 80.51 ± 0.13 & 68.41 ± 0.39 & 84.28 ± 1.73 \\
        &RFS \cite{tian2020rethinking} & ECCV'20 & ResNet-12 & 12.4M & 62.02 ± 0.63 & 79.64 ± 0.44 & 69.74 ± 0.72 & 84.41 ± 0.55 \\
        &FEAT \cite{ye2020FEAT} & CVPR'20 & ResNet-12 & 12.4M & 66.78 ± 0.20 & 82.05 ± 0.14 & 70.80 ± 0.23 & 84.79 ± 0.16\\
        &DeepEMD \cite{zhang2022DeepEMD} & CVPR'20 & ResNet-12 & 12.4M & 65.91 ± 0.82 & 82.41 ± 0.56 & 71.16 ± 0.87 & 86.03 ± 0.58 \\
        &FRN \cite{wertheimer2021FRN} & CVPR'21 & ResNet-12 & 12.4M & 66.45 ± 0.19 & 82.83 ± 0.13 & 71.16 ± 0.22 & 86.01 ± 0.15 \\
        &RENet \cite{kang2021REnet} & ICCV'21 & ResNet-12 & 12.4M & 67.60 ± 0.44 & 82.58 ± 0.30 & 71.61 ± 0.51 & 85.28 ± 0.35 \\
        &Meta-Baseline \cite{chen2021meta} & ICCV'21 & ResNet-12 & 12.4M & 63.17 ± 0.23 & 79.26 ± 0.17 & 68.62 ± 0.27 & 83.29 ± 0.18 \\
        &IE \cite{rizve2021IE} & CVPR'21 & ResNet-12 & 12.4M & 67.28 ± 0.80 & 84.78 ± 0.52 & 72.21 ± 0.90 & 87.08 ± 0.58 \\
        &TPMN \cite{wu2021TPMN} & ICCV'21 & ResNet-12 & 12.4M & 67.64 ± 0.63 & 83.44 ± 0.43 & 72.24 ± 0.70 & 86.55 ± 0.63 \\
        &PSST \cite{chen2021PSST} & CVPR'21 & WRN-28-10 & 36.5M & 74.51 ± 0.46 & 86.60 ± 0.32 & - & - \\
        &OM \cite{qi2021OM} & ICCV'21 & WRN-28-10 & 36.5M & 66.78 ± 0.30 & 85.29 ± 0.41 & 71.54 ± 0.29 & 87.79 ± 0.46 \\
        &Meta-QDA \cite{zhang2021Meta_QDA} & ICCV'21 & WRN-28-10 & 36.5M & 67.83 ± 0.64 & 84.28 ± 0.69 & 74.33 ± 0.65 & 89.56 ± 0.79 \\
        &DeepBDC \cite{xie2022DeepBDC} & CVPR'22 & ResNet-12 & 12.4M & 67.34 ± 0.43 & 84.46 ± 0.28 & 72.34 ± 0.49 & 87.31 ± 0.32 \\
        &SUN \cite{dong2022self} & ECCV'22 & ViT & 12.5M & 67.80 ± 0.45 & 83.25 ± 0.30 & 72.99 ± 0.50 & 86.74 ± 0.33\\
        &FewTURE \cite{hiller2022Fewtrue} & NeurIPS'22 & Swin-T & 29M & 72.40 ± 0.78 & 86.38 ± 0.49 & 76.32 ± 0.87 & 89.96 ± 0.55 \\
        &FGFL \cite{cheng2023FGFL} & ICCV'23 & ResNet-12 & 12.4M & 69.14 ± 0.80 & 86.01 ± 0.62 & 73.21 ± 0.88 & 87.21 ± 0.61 \\
        &CPEA \cite{hao2023CPEA} & ICCV'23 & ViT-S & 22M & 71.97 ± 0.65 & 87.06 ± 0.38 & 76.93 ± 0.70 & 90.12 ± 0.45 \\ 
        &Meta-AdaM \cite{sun2023meta} & NeurIPS'23 & ResNet-12 & 12.4M & 59.89 ± 0.49 & 77.92 ± 0.43 & 65.31 ± 0.48 & 85.24 ± 0.35 \\
        &SMKD \cite{lin2023SMKD} & CVPR'23 & ViT-S & 22M & \underline{74.28 ± 0.18} & \underline{88.82 ± 0.09} & \underline{78.83 ± 0.20} & \underline{91.02 ± 0.12} \\
        &SAFF \cite{cumplido2025slot} & CVPRW'25 & ViT-S & 22M & 71.51 ± 0.63 & 87.19 ± 0.36 & 74.73 ± 0.74 & 88.97 ± 0.43 \\ 
        \cmidrule(lr){2-9}
        &DAFIC (Ours) & WACV'27 & ViT-S & 22M & \textbf{77.71 ± 0.44} & \textbf{90.35 ± 0.51} & \textbf{80.10 ± 0.40} & \textbf{91.05 ± 0.51} \\
        \bottomrule
    \end{tabular}}
    \label{tab:table_fulllist1}
\end{table*}

\begin{table*}[t]
    \centering
    \scriptsize
    \caption{More comprehensive results (\%) on CIFAR-FS and FC100. The ± shows 95\% confidence intervals. The best and second-best results are shown in \textbf{bold} and \underline{underlined}}
    \resizebox{0.99\linewidth}{!}{
    \begin{tabular}{c c l c c c c c c}
    \toprule
         &\multirow{2}{*}{\textbf{Method}} & \multirow{2}{*}{\textbf{Venue}} & \multirow{2}{*}{\textbf{Backbone}} & \multirow{2}{*}{\textbf{$\approx$ \# Params}} & \multicolumn{2}{c}{\textbf{CIFAR-FS}} & \multicolumn{2}{c}{\textbf{FC100}}\\    \cmidrule(lr){6-7}\cmidrule(lr){8-9}
        &\multirow{2}{*}{} &\multirow{2}{*}{} & \multirow{2}{*}{} & \multirow{2}{*}{} & \textbf{5-way 1-shot} & \textbf{5-way 5-shot}  & \textbf{5-way 1-shot} & \textbf{5-way 5-shot}\\ 
        \midrule
        &ProtoNet \cite{snell2017ProtoNet} & NeurIPS'17 & ResNet-12 & 12.4M & 72.20 ± 0.70 & 83.50 ± 0.50 & 41.54 ± 0.76 & 57.08 ± 0.76\\ 
        &TADAM \cite{oreshkin2018FC100} & NeurIPS'18 & ResNet-12 & 12.4M & - & - & 40.10 ± 0.40 & 56.10 ± 0.40 \\
        &MetaOptNet \cite{lee2019MetaOptNet} & CVPR'19 & ResNet-12 & 12.4M & 72.80 ± 0.70 & 84.30 ± 0.50 & 47.20 ± 0.60 & 55.50 ± 0.60\\
        &RFS \cite{tian2020rethinking} & ECCV'20& ResNet-12 & 12.4M & 71.50 ± 0.80 & 86.00 ± 0.50 & 42.60 ± 0.70 & 59.10 ± 0.60 \\
        &DeepEMD \cite{zhang2022DeepEMD} & CVPR'20 & ResNet-12 & 12.4M & - & - & 46.47 ± 0.78 & 63.22 ± 0.71 \\
        &RENet \cite{kang2021REnet} & ICCV'21 & ResNet-12 & 12.4M & 74.51 ± 0.46 & 86.60 ± 0.32 & - & - \\
        &TPMN \cite{wu2021TPMN} & ICCV'21 & ResNet-12 & 12.4M & 75.50 ± 0.90 & 87.20 ± 0.60 & 46.93 ± 0.71 & 63.26 ± 0.74 \\
        &CC+rot \cite{gidaris2019CCrot} & ICCV'21 & WRN-28-10 & 36.5M & 73.62 ± 0.31 & 86.05 ± 0.22 & - & - \\
        &IE \cite{rizve2021IE} & CVPR'21 & ResNet-12 & 12.4M & 77.87 ± 0.85 & 89.74 ± 0.57 & 47.76 ± 0.77 & 65.30 ± 0.76 \\
        &PSST \cite{chen2021PSST} & CVPR'21 & WRN-28-10 & 36.5M & 77.02 ± 0.38 & 88.45 ± 0.35 & - & - \\
        &Meta-QDA \cite{zhang2021Meta_QDA} & ICCV'21 & WRN-28-10 & 36.5M & 75.83 ± 0.88 & 88.79 ± 0.75 & - & - \\
        &SUN \cite{dong2022self} & ECCV'22 & ViT & 12.5M & 78.37 ± 0.46 & 88.84 ± 0.32 & - & -\\
        &FewTURE \cite{hiller2022Fewtrue} & NeurIPS'22 & Swin-T & 29M & 77.76 ± 0.81 & 88.90 ± 0.59 & 47.68 ± 0.78 & 63.81 ± 0.75 \\
        &CPEA \cite{hao2023CPEA} & ICCV'23 & ViT-S & 22M & 77.82 ± 0.66 & 88.98 ± 0.45 & 47.24 ± 0.58 & 65.02 ± 0.60\\ 
        &Meta-AdaM \cite{sun2023meta} & NeurIPS'23 & ResNet-12 & 12.4M & - & - & 41.12 ± 0.49 & 56.14 ± 0.49 \\
        &SMKD \cite{lin2023SMKD} & CVPR'23 & ViT-S & 22M & \underline{80.08 ± 0.18} & \underline{90.63 ± 0.13} & \underline{50.38 ± 0.16} & \underline{68.37 ± 0.16} \\ 
        &SAFF \cite{cumplido2025slot} & CVPRW'25 & ViT-S & 22M & 78.48 ± 0.67 & 90.30 ± 0.41 & 47.17 ± 0.59 & 66.00 ± 0.57 \\ 
        \cmidrule(lr){2-9}
        &DAFIC (Ours) & WACV'27 & ViT-S & 22M & \textbf{84.30 ± 0.44} & \textbf{92.53 ± 0.29} & \textbf{50.52 ± 0.46} & \textbf{69.42 ± 0.44} \\
        \bottomrule
    \end{tabular}}
    \label{tab:table_fulllist2}
\end{table*}

\clearpage

\section{Additional Ablation Studies of DAFIC}
\label{sec:ablation}

\textbf{Choices of Depth-guided Attention (DA).} 
We explore three strategies for deriving depth-guided attention: (1) a patch-average method that uses the mean depth value within each patch as its attention weight; (2) an MLP-based method that applies a lightweight MLP to predict patch-wise attention independently; and (3) our CNN-based design. As shown in Tab.~\ref{tab:rgbd_attention}, the CNN-based design consistently achieves the best performance in both RGB and RGB-D settings. These results align with our design intuition: the patch-average approach produces fixed and coarse attention weights; the MLP-based method, while locally adaptive, is restricted within each patch; the CNN-based method effectively integrates both local and global geometric cues to generate geometry-salient attention.

\begin{table}[h]
\vspace{-2mm}
\caption{Accuracy (\%) comparison between RGB and RGB-D initialization under different DA designs in the DGA module on the TieredImageNet (5-way 1-shot).}
\label{tab:rgbd_attention}
\centering
\scriptsize
\resizebox{0.60\linewidth}{!}{
    \begin{tabular}{lcc}
    \toprule
    \multirow{2}{*}{\textbf{DA Design}} & \multicolumn{2}{c}{\textbf{Initialization}} \\ 
    \cmidrule(lr){2-3}
     & \textbf{RGB} & \textbf{RGB-D} \\
    \midrule
     Patch-average & 77.89{\scriptsize $\pm$0.47} & 79.15{\scriptsize $\pm$0.41} \\
     MLP-based & 78.21{\scriptsize $\pm$0.49} & 79.43{\scriptsize $\pm$0.52} \\
     CNN-based & 78.86{\scriptsize $\pm$0.51} & 80.10{\scriptsize $\pm$0.40}  \\
    \bottomrule
    \end{tabular}}
\end{table}

\vspace{-1mm}
\noindent\textbf{Importance of SBiM.} Tab.~\ref{tab:dense_matching} compares several matching strategies, with all other DAFIC components kept identical. We compare SBiM with its one-way and top-50\% variants, mutual nearest-neighbor matching (MNN), which also performs bidirectional nearest-neighbor search but retains only reciprocal best-match pairs~\cite{liu2022dmn4}, CPEA-style MLP matching~\cite{hao2023CPEA}, and DN4 local nearest neighbor matching~\cite{Li2019DN4} ($k=3$). Overall, SBiM achieves the best performance across all settings, supporting the effectiveness of its bidirectional query-to-support and support-to-query aggregation over one-way or hard reciprocal matching, as well as its complementarity with DGA and the overall DAFIC framework.

\begin{table}[h]
    \vspace{-1.5mm}
    \centering
    \scriptsize
    \setlength{\tabcolsep}{4pt}
    \caption{Comparison of different matching strategies for the DAFIC framework on MiniImageNet and FC100 datasets.}
    \label{tab:dense_matching}
    \resizebox{0.98\linewidth}{!}{
        \begin{tabular}{lcccc} 
        \toprule
        \multirow{2}{*}{\textbf{Matching Strategy}} & \multicolumn{2}{c}{\textbf{MiniImageNet}} & \multicolumn{2}{c}{\textbf{FC100}} \\ 
        \cmidrule(lr){2-3} \cmidrule(lr){4-5}
        & \textbf{1-shot}  & \textbf{5-shot} & \textbf{1-shot}  & \textbf{5-shot} \\ 
        \midrule
        SBiM & 77.71{\scriptsize $\pm$0.44} & 90.35{\scriptsize $\pm$0.51} & 50.52{\scriptsize $\pm$0.46} & 69.42{\scriptsize $\pm$0.44} \\
        SBiM (top 50\% patches) & 77.02{\scriptsize $\pm$0.42} & 89.92{\scriptsize $\pm$0.40} & 50.15{\scriptsize $\pm$0.47} & 68.15{\scriptsize $\pm$0.43} \\
        SBiM (one-way) & 66.05{\scriptsize $\pm$0.47} & 80.17{\scriptsize $\pm$0.36} & 44.61{\scriptsize $\pm$0.44} & 59.41{\scriptsize $\pm$0.43} \\
        Mutual-NN (MNN) & 76.38{\scriptsize $\pm$0.40} & 89.18{\scriptsize $\pm$0.29} & 49.21{\scriptsize $\pm$0.55} & 67.75{\scriptsize $\pm$0.40}\\
        MLP (CPEA) & 76.20{\scriptsize $\pm$0.41} & 88.79{\scriptsize $\pm$0.38} & 48.56{\scriptsize $\pm$0.60} & 67.05{\scriptsize $\pm$0.59} \\
        DN4 (k=3) & 67.49{\scriptsize $\pm$0.43} & 85.63{\scriptsize $\pm$0.29} & 47.97{\scriptsize $\pm$0.45} & 65.22{\scriptsize $\pm$0.43} \\
        \bottomrule
        \end{tabular}
    }
\end{table}

\vspace{-1mm}
\noindent\textbf{Ablation of $\beta$ in the DA component of DGA module.} Tab.~\ref{tab:ablation_beta} investigates the attention-strength parameter $\beta$ within the depth-guided attention (DA) component of the DGA module on MiniImageNet. We compare fixed values of $\beta$ against making it a learnable parameter with different initializations. The results indicate that setting $\beta$ as a learnable parameter initialized at $5.0$ achieves the best performance. These results suggest that a moderately larger initial attention scale is beneficial, whereas overly large scaling (e.g., $\beta=10.0$) slightly degrades accuracy. An evaluation of the initial scale value may be relevant due to its effect on the adaptive attention learning of our DAFIC framework.

\begin{table}[h]
    \vspace{-1.5mm}
    \centering
    \scriptsize
    \setlength{\tabcolsep}{4pt}
    \caption{Ablation on the attention-strength parameter $\beta$ on MiniImageNet.}
    \label{tab:ablation_beta}
        \begin{tabular}{lccc} 
        \toprule
         \textbf{$\beta$ Setting} &  \textbf{Learnable} & \textbf{1-shot}  & \textbf{5-shot} \\ 
        \midrule
         Fixed $\beta=1.0$ & No & 76.62{\scriptsize $\pm$0.44} & 89.59{\scriptsize $\pm$0.51} \\
         Fixed $\beta=5.0$ & No & 77.02{\scriptsize $\pm$0.42} & 89.92{\scriptsize $\pm$0.40} \\
         Init $\beta=1.0$ & Yes & 76.68{\scriptsize $\pm$0.51} & 89.66{\scriptsize $\pm$0.36} \\
         Init $\beta=5.0$ & Yes & 77.71{\scriptsize $\pm$0.44} & 90.35{\scriptsize $\pm$0.51} \\
         Init $\beta=10.0$ & Yes & 77.34{\scriptsize $\pm$0.40} & 90.22{\scriptsize $\pm$0.38} \\
        \bottomrule
        \end{tabular}
\end{table}

\vspace{-1mm}
\noindent\textbf{Ablation of $\lambda$ in the GA component of DGA module.} Tab.~\ref{tab:ablation_lambda} ablates the magnitude parameter $\lambda$ within the geometry-aware adaptation (GA) component of the DGA module on MiniImageNet. As mentioned in the main manuscript, we adopt $\lambda=2$ following prior works \cite{hao2023CPEA, jiang2024CPES, cumplido2025slot}. Our empirical results also show that this setting gives the best performance among the values considered.

\begin{table}[h]
    \vspace{-1.5mm}
    \centering
   \scriptsize
    \setlength{\tabcolsep}{4pt}
    \caption{Ablation on the magnitude $\lambda$ on MiniImageNet.}
    \label{tab:ablation_lambda}
        \begin{tabular}{lcc} 
        \toprule
         \textbf{$\lambda$} & \textbf{1-shot}  & \textbf{5-shot} \\ 
        \midrule
         $1$ & 76.97{\scriptsize $\pm$0.46} & 89.50{\scriptsize $\pm$0.44} \\
         $2$ & 77.71{\scriptsize $\pm$0.44} & 90.35{\scriptsize $\pm$0.51} \\
         $4$ & 77.18{\scriptsize $\pm$0.44} & 89.77{\scriptsize $\pm$0.43} \\
         $8$ & 76.99{\scriptsize $\pm$0.60} & 89.61{\scriptsize $\pm$0.59} \\
        \bottomrule
        \end{tabular}
\end{table}

\vspace{-1mm}
\noindent\textbf{Ablation of $\alpha$ and $\gamma$ for the final loss.} Tab.~\ref{tab:ablation_alpha_gamma} investigates the loss weights $\alpha$ (for local matching) and $\gamma$ (for auxiliary global supervision) on MiniImageNet. By fixing the primary loss weight $\alpha$ at 1.0, we evaluate the impact of the auxiliary weight $\gamma$. Empirical results show that setting $\gamma = 0.5$ yields the best performance, indicating that moderate auxiliary global supervision is beneficial for DAFIC.

\begin{table}[h]
    \vspace{-1.5mm}
    \centering
    \scriptsize
    \setlength{\tabcolsep}{4pt}
    \caption{Ablation on loss weights $\alpha$ for local matching and $\gamma$ for auxiliary global supervision on MiniImageNet.}
    \label{tab:ablation_alpha_gamma}
        \begin{tabular}{lccc} 
        \toprule
         \textbf{$\alpha$} & \textbf{$\gamma$} & \textbf{1-shot}  & \textbf{5-shot} \\ 
        \midrule
         1.0 & 0 & 77.42{\scriptsize $\pm$0.49} & 90.14{\scriptsize $\pm$0.41} \\
         1.0 & 0.5 & 77.71{\scriptsize $\pm$0.41} & 90.35{\scriptsize $\pm$0.51} \\
         1.0 & 1.0 & 77.60{\scriptsize $\pm$0.44} & 90.24{\scriptsize $\pm$0.46} \\
        \bottomrule
        \end{tabular}
\end{table}

\vspace{-1mm}
\noindent\textbf{Independent contributions of DGA and SBiM.}
Tab.~\ref{tab:DGASBiM} evaluates DGA and SBiM without RGB-D pre-training to isolate their contributions. Without FSL adaptation, SBiM already outperforms MLP, showing the benefit of bidirectional dense matching. Under CPEA-style adaptation, MLP performs better than SBiM, suggesting that uniform global-context adaptation is better aligned with the original CPEA classifier. In contrast, DGA consistently benefits both classifiers, and DGA+SBiM achieves the best performance. This supports our design in which DGA performs geometry-guided adaptation, while SBiM effectively exploits the refined features through bidirectional support-query matching.

\begin{table}[h]
\centering
\scriptsize
\setlength{\tabcolsep}{5.0pt}
\renewcommand{\arraystretch}{1.05}
\caption{Effect of DGA and SBiM \emph{without RGB-D pre-training} on MiniImageNet. Results are 5-way 1-/5-shot accuracies (\%).}
\label{tab:DGASBiM}
\begin{tabular}{lcc}
\toprule
\textbf{FSL Adaptation} & \textbf{MLP} & \textbf{SBiM} \\
\midrule
None       & 30.49 / 41.23 & 32.93 / 48.10 \\
CPEA-style & 31.01 / 46.66 & 29.73 / 44.78 \\
DGA        & 32.61 / 49.67 & 34.36 / 52.81 \\
\bottomrule
\end{tabular}
\end{table}

\vspace{-1mm}
\noindent\textbf{Variance over multiple training runs.}
Table~\ref{tab:seed_variance} reports DAFIC performance across three runs with different random seeds. Results are reported as mean $\pm$ standard deviation across runs, rather than episodic confidence intervals. This indicates relatively stable training across datasets and 5-way 1-/5-shot settings.
\vspace{2mm}

\begin{table}[h]
\centering
\scriptsize
\setlength{\tabcolsep}{4.5pt}
\renewcommand{\arraystretch}{1.05}
\caption{Training stability of DAFIC across three runs with different random seeds. Results are reported as mean $\pm$ standard deviation (\%) across runs.}
\label{tab:seed_variance}
\begin{tabular}{lcc}
\toprule
\textbf{Dataset} & \textbf{1-shot} & \textbf{5-shot} \\
\midrule
MiniImageNet   & 77.42 $\pm$ 0.26 & 90.29 $\pm$ 0.15 \\
TieredImageNet & 80.09 $\pm$ 0.17 & 90.94 $\pm$ 0.15 \\
CIFAR-FS       & 84.45 $\pm$ 0.16 & 92.40 $\pm$ 0.16 \\
FC100          & 50.36 $\pm$ 0.21 & 69.37 $\pm$ 0.32 \\
\bottomrule
\end{tabular}
\end{table}

\noindent\textbf{Latency comparison.}
Table~\ref{tab:latency} extends the latency analysis in Main Tab.5 by reporting the full inference latency with and without cached depth maps, comparing DAFIC with the CPEA baseline and SMKD, the strongest-performing competing method in our comparison. In our standard setting, depth maps are precomputed offline, so DAFIC introduces no online depth-estimation overhead and achieves a latency of only 1.39\,ms per image, slightly lower than CPEA (1.43\,ms). Even when depth is estimated on-the-fly using Depth Anything V2-ViT/B~\cite{yang2024depth2}, the total latency is 74.02\,ms per image, remaining lower than SMKD (78.40\,ms). These results show that DAFIC preserves efficient inference with cached depth while remaining competitive in latency when online depth estimation is included.

\vspace{-1mm}
\begin{table}[h]
\centering
\scriptsize
\setlength{\tabcolsep}{4.0pt}
\renewcommand{\arraystretch}{1.05}
\caption{Inference latency with and without cached depth maps. Cached depth maps are precomputed offline, whereas on-the-fly inference includes Depth Anything V2-B.}
\label{tab:latency}
\begin{tabular}{lccc}
\toprule
\textbf{Method} &
\textbf{FSIC Latency} &
\textbf{Depth Latency} &
\textbf{Total Latency} \\
\midrule
CPEA                     & 1.43  & --    & 1.43 \\
DAFIC (cached)           & 1.39  & --    & 1.39 \\
DAFIC (on-the-fly)       & 1.39  & 72.63 & 74.02 \\
SMKD                     & 78.40 & --    & 78.40 \\
\bottomrule
\end{tabular}
\end{table}

\noindent\textbf{Sensitivity to depth-prior quality.}
To further examine whether DAFIC depends on a specific external depth estimator, we replace Depth Anything V2~\cite{yang2024depth2} with a weaker alternative depth prior generated by a DPT head with a DINO-v2 ViT-Small backbone~\cite{oquab2024dinov2}. Qualitative examples are shown in Fig.~\ref{fig:lowdepth}. The resulting depth maps are used consistently throughout RGB-D pre-training, episodic meta-training, and evaluation. As shown in Tab.~\ref{tab:depth_degrade_train}, performance decreases from 77.71/90.35\% to 76.16/89.05\% on MiniImageNet for 5-way 1-/5- shot, while remaining above the CPEA baseline by 3.91/1.81\%. This indicates that depth quality affects performance, while DAFIC remains effective with a weaker geometric prior, consistent with our design of using depth as a soft structural cue rather than relying on a specific depth estimator.

\begin{table}[h] 
    \centering
    \makeatletter\def\@captype{table}\makeatother 
    \caption{Sensitivity of DAFIC to an alternative weaker depth prior throughout training and evaluation on MiniImageNet.}
    \vspace{-1mm}
    \scriptsize
    \label{tab:depth_degrade_train}
    \setlength{\tabcolsep}{6pt}
    \resizebox{0.95\linewidth}{!}{
        \begin{tabular}{l c c c}
            \toprule
            \textbf{Method} & \textbf{Depth Source} & \textbf{1-shot} & \textbf{5-shot} \\
            \midrule
            CPEA  & --                    & 72.25$\pm$0.65 & 87.24$\pm$0.38 \\
            DAFIC & Depth Anything V2     & 77.71$\pm$0.44 & 90.35$\pm$0.51 \\
            DAFIC & DINOv2-S + DPT        & 76.16$\pm$0.40 & 89.05$\pm$0.22 \\
            \bottomrule
        \end{tabular}
    }
    \vspace{-4mm}
\end{table}

\begin{figure}[h]
    \centering
    \includegraphics[width=0.40\textwidth]{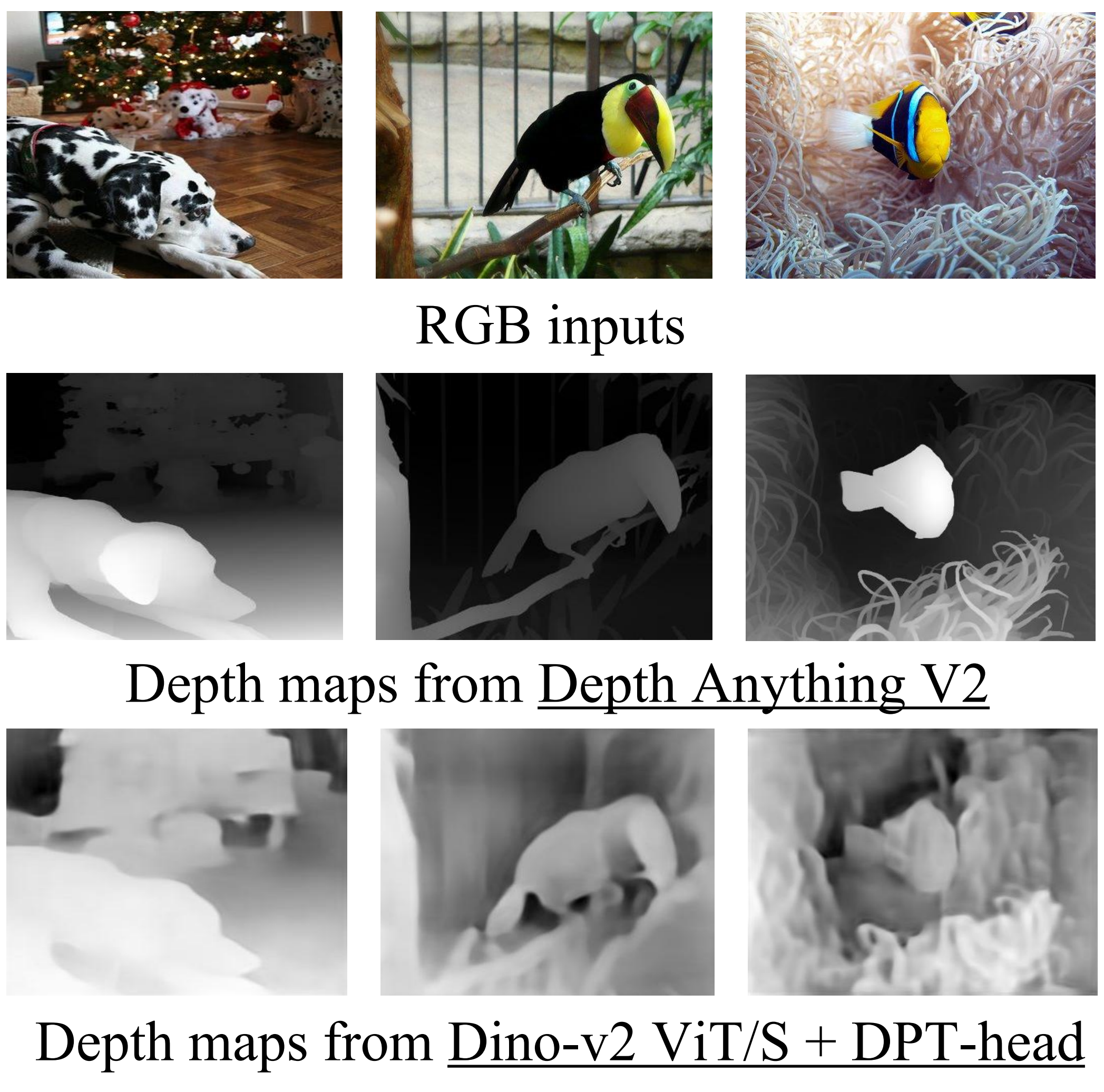}
    \caption{Qualitative examples of depth maps generated by Depth Anything V2 (second row) and the weaker DINO-v2 ViT/S + DPT (third row) depth estimator.}
    \label{fig:lowdepth}
\end{figure}

\vspace{-4mm}
\section{Discussion of DAFIC}
\label{sec:discussion}
We provide additional discussions of DAFIC on its novelty, limitations, and potential future directions. 

\noindent\textbf{Innovation beyond existing patch-level FSIC methods.}
The novelty of DAFIC does not lie in the individual use of monocular depth estimation or bidirectional matching, nor simply in adding depth as an extra input modality. Instead, DAFIC introduces a unified geometry-guided FSIC formulation that integrates geometric information into representation learning, episodic patch adaptation, and dense support-query matching to address \emph{appearance ambiguity} and \emph{background clutter} from RGB-only FSIC scenarios.

This formulation differs from existing patch-level FSIC methods in three aspects: 1) \textbf{Geometry-aware representation learning.} RGB-D self-supervised pre-training incorporates frozen and task-agnostic monocular depth into the backbone, producing geometry-aware patch representations rather than relying solely on appearance-based features.
2) \textbf{Geometry-conditioned episodic adaptation.} Unlike CPEA-style adaptation that injects class-level information into patch representations without explicitly considering geometric relevance, DGA uses depth-derived geometry to selectively preserve local cues for geometry-salient patches while incorporating geometry-enriched class context, and steers background regions toward a global representation. This produces geometry-refined patch sets that are less sensitive to background-dominated or semantically misaligned regions.
3) \textbf{Bidirectional matching of geometry-refined features.} We regard FSIC as a patch-matching problem and design SBiM to perform bidirectional support-query matching on the DGA-refined features, reducing spurious correspondences during dense comparison.

These components form a coherent pipeline rather than independent additions. Thus, the contribution of DAFIC lies in this geometry-guided formulation and, in particular, the coupling of geometry-conditioned episodic adaptation mechanism with bidirectional set matching, rather than simply combining existing components.

\noindent\textbf{Segmentation methods for future work.}
Segmentation-based methods, such as SAM~\cite{kirillov2023sam,carion2026sam3} and other open-world models~\cite{ren2024groundedsam}, indeed provide a promising complementary direction for our future work. However, segmentation also has inherent limitations for FSIC that need to be systematically explored: 1) Segmentation-based approaches typically require an additional prompt or a mask-selection mechanism to identify which masks are class-relevant for each few-shot task; 2) Foreground segmentation can suppress or remove background, but it does not by itself resolve few-shot matching ambiguities, such as intra-object appearance similarity, part-level misalignment, and support-query patch correspondence. 3) In many FSIC cases, object parts, shape cues, and even surrounding regions can be informative, while a hard segmentation mask may inadvertently remove such useful cues. In comparison, depth provides a softer and task-agnostic alternative. Our DAFIC incorporates geometric cues into representation learning and leverages them as a soft attention prior, together with geometry-enriched class context, to improve local-to-global semantic alignment and support SBiM for robust patch matching. Therefore, how to effectively address the above concerns and explore segmentation-based paradigms for FSIC remains an important direction for future work.

\noindent\textbf{Foundation models and the strict FSIC protocol.}
Modern foundation representations provide an important reference for FSIC. The key difference is how this external information enters the FSIC pipeline: foundation encoders directly provide externally pre-trained semantic representations, whereas DAFIC uses a learnable backbone whose parameters are optimized only on the standard meta-train split, with fixed, task-agnostic depth predictions serving as auxiliary geometric inputs. As a diagnostic, we evaluate frozen DINOv2~\cite{oquab2024dinov2} and CLIP~\cite{radford2021learning} visual representations under the same prototypical evaluation protocol. DINOv2 achieves 94.90/98.59\% on MiniImageNet and 93.97/97.64\% on TieredImageNet, while CLIP achieves 85.57/95.64\% on MiniImageNet for 1-/5-shot. These results highlight the strong few-shot capability enabled by large-scale external semantic pre-training, even with frozen encoders and a simple prototypical classifier.

\section{Effect of Depth Cues on FSIC Methods}
\label{sec:depth_potential}

To explicitly account for the external depth prior used by DAFIC, we construct matched-prior controls by augmenting representative FSIC methods with the same estimated depth maps. Specifically, depth is concatenated with RGB as an additional input channel, while the original backbone, training protocol, and evaluation setting of each method are otherwise kept unchanged. All depth maps are generated offline using the same fixed monocular depth estimator. Adding the depth channel changes only the input projection parameters marginally relative to the full model size.

As shown in Tables~\ref{tab:table_depth1} and \ref{tab:table_depth2}, providing the same depth prior improves most evaluated methods across MiniImageNet, TieredImageNet, CIFAR-FS, and FC100. This confirms that externally estimated depth itself provides a useful additional signal and accounts for part of the improvement observed in depth-assisted FSIC. Importantly, however, DAFIC remains above the directly matched CPEA+Depth baseline on all four benchmarks, by +2.51/+2.06, +2.06/+0.94, +2.65/+1.98, and +2.46/+2.77 percentage points on MiniImageNet, TieredImageNet, CIFAR-FS, and FC100, respectively, for 5-way 1-/5-shot classification. Thus, access to the same external depth prior alone does not explain the full performance gain of DAFIC.

Table~\ref{tab:table_depth3} further extends this analysis to the fine-grained CUB~\cite{welinder2010caltech} and StanfordCars~\cite{krause2013StanfordCars} datasets. Depth again benefits many methods, although it is not uniformly advantageous in every setting. DAFIC achieves 85.94/93.67\% on CUB and 93.52/96.85\% on StanfordCars, remaining above CPEA+Depth by +2.83/+1.82 and +1.16/+1.01 percentage points, respectively. Overall, these matched controls show that depth is a valuable complementary cue for FSIC, while the remaining gains of DAFIC support the benefit of learning how to integrate and exploit this geometric prior rather than merely providing an additional depth channel.

\begin{table*}[t]
    \centering
    \scriptsize
    \setlength{\tabcolsep}{3pt} 
    \caption{Results (\%) of FSIC methods with RGB-D inputs on MiniImageNet and TieredImageNet. The $\pm$ denotes 95\% confidence intervals. Adding depth as an extra input channel only marginally increases the input projection parameters relative to the full model size. For FewTURE, which provides both Swin-T and ViT-S variants, we report the ViT-S results for a backbone-matched comparison.}
    \resizebox{0.80\linewidth}{!}{
    \begin{tabular}{c l c c c c c c c}
    \toprule
         &\multirow{2}{*}{\textbf{Method}} & \multirow{2}{*}{\textbf{Backbone}} & \multirow{2}{*}{\textbf{$\approx$ \# Params}} & \multicolumn{2}{c}{\textbf{MiniImageNet}} & \multicolumn{2}{c}{\textbf{TieredImageNet}}\\ 
         \cmidrule(lr){5-6}\cmidrule(lr){7-8}
        &\multirow{2}{*}{} & \multirow{2}{*}{} & \multirow{2}{*}{} & \textbf{5-way 1-shot} & \textbf{5-way 5-shot}  & \textbf{5-way 1-shot} & \textbf{5-way 5-shot}\\ 
        \midrule
        &ProtoNet \cite{snell2017ProtoNet}& ResNet-12 & 12.4M & 62.29 ± 0.33 & 79.46 ± 0.48 & 68.25 ± 0.23 & 84.01 ± 0.56 \\
        &ProtoNet+Depth & ResNet-12 & 12.4M &  67.56 ± 0.21 &  84.11 ± 0.14 &  72.62 ± 0.42 &  86.88 ± 0.16 \\
        \hline
        &DN4 \cite{Li2019DN4} & ResNet-12 & 12.4M & 56.93 ± 0.62 & 73.97 ± 0.49 & 66.25 ± 0.75 & 82.62 ± 0.51 \\
        &DN4+Depth & ResNet-12 & 12.4M & 65.27 ± 0.64 & 82.33 ± 0.40 & 70.83 ± 0.73 & 86.67 ± 0.47 \\
        \hline
        &CAN \cite{hou2019cross} & ResNet-12 & 12.4M & 63.85 ± 0.48 & 79.44 ± 0.34 & 69.89 ± 0.51 & 84.23 ± 0.37 \\
        &CAN+Depth & ResNet-12 & 12.4M & 70.85 ± 0.48 & 84.81 ± 0.32 & 75.77 ± 0.50 & 87.61 ± 0.37 \\
        \hline
        &RFS \cite{tian2020rethinking} & ResNet-12 & 12.4M & 62.02 ± 0.63 & 79.64 ± 0.44 & 69.74 ± 0.72 & 84.41 ± 0.55 \\
        &RFS+Depth & ResNet-12 & 12.4M & 70.03 ± 0.60 & 85.22 ± 0.44 & 75.05 ± 0.72 & 87.98 ± 0.53 \\
        \hline
        &FEAT \cite{ye2020FEAT} & ResNet-12 & 12.4M & 66.78 ± 0.20 & 82.05 ± 0.14 & 70.80 ± 0.23 & 84.79 ± 0.16\\
        &FEAT+Depth & ResNet-12 & 12.4M & 70.92 ± 0.20 & 84.97 ± 0.13 & 73.81 ± 0.23 & 86.61 ± 0.15\\
        \hline
        &Meta-Baseline \cite{chen2021meta} & ResNet-12 & 12.4M & 63.17 ± 0.23 & 79.26 ± 0.17 & 68.62 ± 0.27 & 83.29 ± 0.18 \\
        &Meta-Baseline+Depth & ResNet-12 & 12.4M & 68.40 ± 0.52 & 84.10 ± 0.34 & 73.21 ± 0.49 & 87.66 ± 0.25  \\
        \hline
        &FRN \cite{wertheimer2021FRN} & ResNet-12 & 12.4M & 66.45 ± 0.19 & 82.83 ± 0.13 & 71.16 ± 0.22 & 86.01 ± 0.15 \\
        &FRN+Depth & ResNet-12 &  12.4M & 71.61 ± 0.19   &  84.99 ± 0.13  &  75.36 ± 0.22  & 88.38 ± 0.15 \\
        \hline
        &RENet \cite{kang2021REnet} & ResNet-12 & 12.4M & 67.60 ± 0.44 & 82.58 ± 0.30 & 71.61 ± 0.51 & 85.28 ± 0.35 \\
        &RENet+Depth & ResNet-12 & 12.4M & 72.68 ± 0.44  & 86.82 ± 0.29 & 75.19 ± 0.46 & 88.96 ± 0.35 \\
        \hline
        &SUN \cite{dong2022self} & ViT-S & 22M & 67.80 ± 0.45 & 83.25 ± 0.30 & 72.99 ± 0.50 & 86.74 ± 0.33\\
        &SUN+Depth & ViT-S & 22M & 72.08 ± 0.43 & 85.53 ± 0.30 & 75.81 ± 0.49 & 88.92 ± 0.35\\
        \hline
        &FewTURE \cite{hiller2022Fewtrue} & ViT-S & 22M & 68.02 ± 0.88 & 84.51 ± 0.53 & 72.96 ± 0.92 & 86.43 ± 0.67 \\
        &FewTURE+Depth & ViT-S & 22M & 73.34 ± 0.81 & 86.55 ± 0.49 & 76.75 ± 0.92 & 88.78 ± 0.62 \\
        \hline
        &SMKD \cite{lin2023SMKD} & ViT-S & 22M & 74.28 ± 0.18 & 88.82 ± 0.09 & 78.83 ± 0.20 & 91.02 ± 0.12 \\
        &SMKD+Depth & ViT-S & 22M &  75.87 ± 0.18 & 89.36 ± 0.10 &  79.52 ± 0.16  & 91.33 ± 0.20  \\
        \hline
        &CPEA \cite{hao2023CPEA} & ViT-S & 22M & 72.25 ± 0.65 & 87.24 ± 0.38 & 77.12 ± 0.70 & 90.12 ± 0.45 \\ 
        &CPEA+Depth & ViT-S & 22M & 75.20 ± 0.41 & 88.29 ± 0.38 & 78.04 ± 0.70 & 90.11 ± 0.41 \\
        \bottomrule    
    \end{tabular}}
    \label{tab:table_depth1}
\end{table*}

\begin{table*}[t]
    \centering\
    \scriptsize
    \setlength{\tabcolsep}{3pt} 
    \renewcommand{\arraystretch}{0.98} 
    \caption{FSIC method with RGB-D input results (\%) on CIFAR-FS and FC100. The ± shows 95\% confidence intervals.}
    \resizebox{0.80\linewidth}{!}{
    \begin{tabular}{c l c c c c c c c}
    \toprule
         &\multirow{2}{*}{\textbf{Method}} & \multirow{2}{*}{\textbf{Backbone}} & \multirow{2}*{\textbf{$\approx$ \# Params}} & \multicolumn{2}{c}{\textbf{CIFAR-FS}} & \multicolumn{2}{c}{\textbf{FC100}}\\ 
         \cmidrule(lr){5-6}\cmidrule(lr){7-8}
        &\multirow{2}{*}{} & \multirow{2}{*}{} & \multirow{2}{*}{} & \textbf{5-way 1-shot} & \textbf{5-way 5-shot}  & \textbf{5-way 1-shot} & \textbf{5-way 5-shot}\\ 
        \midrule
        &ProtoNet \cite{snell2017ProtoNet} & ResNet-12 & 12.4M & 72.20 ± 0.70 & 83.50 ± 0.50 & 41.54 ± 0.76 & 57.08 ± 0.76\\ 
        &ProtoNet+Depth  & ResNet-12 & 12.4M & 76.17 ± 0.66 & 86.92 ± 0.44 &  43.77 ± 0.62 & 60.22 ± 0.74\\ 
        \hline
        &DN4 \cite{Li2019DN4} & ResNet-12 & 12.4M & 71.88 ± 0.68 & 85.16 ± 0.49 & 39.56 ± 0.56 & 56.50 ± 0.54\\ 
        &DN4+Depth & ResNet-12 & 12.4M & 76.23 ± 0.68 & 88.68 ± 0.41 & 41.47 ± 0.57 & 59.21 ± 0.55 \\
        \hline
        &RFS \cite{tian2020rethinking} & ResNet-12 & 12.4M & 71.50 ± 0.80 & 86.00 ± 0.50 & 42.60 ± 0.70 & 59.10 ± 0.60 \\
        &RFS+Depth & ResNet-12 & 12.4M & 75.66 ± 0.74 & 87.94 ± 0.48 & 43.85 ± 0.61 & 60.78 ± 0.59 \\
        \hline
        &RENet \cite{kang2021REnet} & ResNet-12 & 12.4M & 74.51 ± 0.46 & 86.60 ± 0.32 & 44.61 ± 0.46 & 60.95 ± 0.41 \\
        &RENet+Depth & ResNet-12 & 12.4M & 78.57 ± 0.44 & 90.23 ± 0.28 & 45.26 ± 0.41 & 61.67 ± 0.39 \\
        \hline
        &SUN \cite{dong2022self} & ViT-S & 22M & 78.37 ± 0.46 & 88.84 ± 0.32 & 45.68 ± 0.52 & 61.45 ± 0.66 \\
        &SUN+Depth & ViT-S & 22M & 81.75 ± 0.49 & 91.25 ± 0.44 & 46.55 ± 0.58 & 62.31 ± 0.61 \\
        \hline
        &FewTURE \cite{hiller2022Fewtrue} & ViT-S & 22M & 76.10 ± 0.88 & 86.14 ± 0.64 & 46.20 ± 0.79 & 63.14 ± 0.73 \\
        &FewTURE+Depth & ViT-S & 22M & 80.27 ± 0.82 & 90.25 ± 0.58 & 47.42 ± 0.84 & 64.22 ± 0.73 \\
        \hline
        &SMKD \cite{lin2023SMKD} & ViT-S & 22M & 80.08 ± 0.18 & 90.63 ± 0.13 & 50.38 ± 0.16 & 68.37 ± 0.16 \\
        &SMKD+Depth & ViT-S & 22M & 81.20 ± 0.18 & 91.21 ± 0.13 & 51.77 ± 0.17 & 69.63 ± 0.16 \\
        \hline
        &CPEA \cite{hao2023CPEA} & ViT-S & 22M & 78.27 ± 0.66 & 89.30 ± 0.45 & 46.51 ± 0.58 & 65.18 ± 0.60\\
        &CPEA+Depth & ViT-S & 22M & 81.65 ± 0.60 & 90.55 ± 0.41 & 48.06 ± 0.60 & 66.65 ± 0.59\\
        \bottomrule
    \end{tabular}}
    \label{tab:table_depth2}
\end{table*}

\clearpage

\begin{table*}[!h]
    \centering
    \scriptsize
    \setlength{\tabcolsep}{3pt} 
    \renewcommand{\arraystretch}{0.98} 
    \caption{FSIC method with RGB-D input results (\%) on CUB and StanfordCars. The ± shows 95\% confidence intervals.}
    \resizebox{0.80\linewidth}{!}{
    \begin{tabular}{c l c c c c c c c}
    \toprule
         &\multirow{2}{*}{\textbf{Method}} & \multirow{2}{*}{\textbf{Backbone}} & \multirow{2}*{\textbf{$\approx$ \# Params}} & \multicolumn{2}{c}{\textbf{CUB}} & \multicolumn{2}{c}{\textbf{StanfordCars}}\\ 
         \cmidrule(lr){5-6}\cmidrule(lr){7-8}
        &\multirow{2}{*}{} & \multirow{2}{*}{} & \multirow{2}{*}{} & \textbf{5-way 1-shot} & \textbf{5-way 5-shot}  & \textbf{5-way 1-shot} & \textbf{5-way 5-shot}\\ 
        \midrule
        &ProtoNet \cite{snell2017ProtoNet} & ResNet-12 & 12.4M & 66.29 ± 0.92 & 82.58 ± 0.58 & 70.07 ± 0.18 &  84.21 ± 0.35\\ 
        &ProtoNet+Depth  & ResNet-12 & 12.4M & 68.50 ± 0.52 & 83.40 ± 0.22 & 73.21 ± 0.11 &  85.54 ± 0.20\\ 
        \hline
        &DN4 \cite{Li2019DN4} & ResNet-12 & 12.4M & 79.01 ± 0.64 & 90.03 ± 0.35 & 86.08 ± 0.59 & 95.35 ± 0.59\\ 
        &DN4+Depth & ResNet-12 & 12.4M & 81.67 ± 0.64 & 92.18 ± 0.33 & 87.21 ± 0.59 & 96.26 ± 0.59 \\
        \hline
        &FEAT \cite{ye2020FEAT} & ResNet-12 & 12.4M & 82.14 ± 0.19 & 92.05 ± 0.38  & 86.58 ± 0.65 & 95.13 ± 0.28\\
        &FEAT+Depth & ResNet-12 & 12.4M & 81.97 ± 0.19 & 91.56 ± 0.11 & 87.54 ± 0.17 & 94.68 ± 0.10 \\
        \hline
        &RENet \cite{kang2021REnet} & ResNet-12 & 12.4M & 79.24 ± 0.43 & 90.51 ± 0.25 & 86.71 ± 0.40 & 94.45 ± 0.21 \\
        &RENet+Depth & ResNet-12 & 12.4M & 79.52 ± 0.44 & 90.89 ± 0.25 & 91.21 ± 0.39 & 96.75 ± 0.21\\
        \hline
        &FewTURE \cite{hiller2022Fewtrue} & ViT-S & 22M & 58.36 ± 0.95 & 69.51 ± 0.73 & 31.62 ± 0.66 & 49.41 ± 0.74 \\
        &FewTURE+Depth & ViT-S & 22M & 77.91 ± 0.91 & 87.45 ± 0.58 & 85.24 ± 0.77 & 91.13 ± 0.52 \\
        \hline
        &SMKD \cite{lin2023SMKD} & ViT-S & 22M & 79.62 ± 0.19 & 90.78 ± 0.11  & 83.90 ± 0.20 &  93.98 ± 0.11\\
        &SMKD+Depth & ViT-S & 22M & 80.55 ± 0.20  & 91.02 ± 0.12  & 84.50 ± 0.21 & 94.22 ± 0.11 \\
        \hline
        &CPEA \cite{hao2023CPEA} & ViT-S & 22M & 65.12 ± 0.72  & 81.05 ± 0.50 & 35.47 ± 0.59 & 64.93 ± 0.59\\
        &CPEA+Depth & ViT-S & 22M & 83.11 ± 0.60  & 91.85 ± 0.34 & 92.36 ± 0.47 & 95.84 ± 0.29\\
        \hline
        &DAFIC (Ours) & ViT-S & 22M & 85.94 ± 0.63  & 93.67 ± 0.32 & 93.52 ± 0.47 & 96.85 ± 0.25\\
        \bottomrule
    \end{tabular}}
    \label{tab:table_depth3}
\end{table*}
\clearpage
\end{document}